\documentclass{article}
\usepackage{iclr2027_conference,times}
\iclrfinalcopy

\usepackage{amsmath,amsfonts,bm}

\def\eqref#1{equation~\ref{#1}}

\def\1{\bm{1}}

\def\vzero{{\bm{0}}}

\def\vx{{\bm{x}}}

\DeclareMathAlphabet{\mathsfit}{\encodingdefault}{\sfdefault}{m}{sl}
\SetMathAlphabet{\mathsfit}{bold}{\encodingdefault}{\sfdefault}{bx}{n}

\newcommand{\E}{\mathbb{E}}
\newcommand{\Ls}{\mathcal{L}}

\usepackage{hyperref}
\usepackage{url}
\usepackage{booktabs}
\usepackage{graphicx}
\usepackage{placeins}
\usepackage{amsmath,amssymb}
\usepackage{xcolor}
\usepackage{makecell}
\usepackage{arydshln}
\usepackage{pifont}
\newcommand{\cmark}{\ding{51}} 
\newcommand{\xmark}{\ding{55}} 

\title{Multimodal Co-Generation with Continuous Bitstream Diffusion}
\newcommand{\authorentry}[2]{\textbf{#1}\newline\mdseries #2}
\author{%
\begin{tabular}[t]{@{}p{0.30\textwidth}p{0.30\textwidth}p{0.30\textwidth}@{}}
\authorentry{Georgios Batzolis$^*$}{Devotion AI Labs} &
\authorentry{Georgios Kontos$^*$}{University of Cambridge} &
\authorentry{Bruno Trentini$^*$}{University of Oxford \newline NVIDIA} \\[10pt]
\authorentry{Pranam Chatterjee}{University of Pennsylvania} &
\authorentry{Micha Livne}{NVIDIA} &
\authorentry{Mark Girolami}{Devotion AI Labs \newline University of Cambridge} \\[10pt]
& \authorentry{Luca Ambrogioni}{Radboud University} &
\end{tabular}%
}

\begin{document}

\maketitle
\lhead{Preprint}

\begingroup
\makeatletter
\long\def\@makefntext#1{\noindent#1}
\makeatother
\renewcommand{\thefootnote}{}
\footnotetext[0]{%
$^*$Equal contribution.
}
\endgroup

\begin{abstract}
We show that continuous bitstream diffusion (CoBit), originally developed for language modeling, provides a unified framework for joint and conditional multimodal generation using the same training and sampling protocol. Across image--text, speech--text and protein sequence--structure modeling, models are trained separately from scratch, but each uses a single input projection, transformer backbone, bitwise prediction head and binary score-matching objective shared across its modalities. Within each setting, a single checkpoint supports joint generation and the conditional directions included in training by changing which parts of the bitstream are observed. We first validate the approach on MNIST-Sum, a controlled image--text task pairing raw binary images with equations encoded as text, achieving near-perfect conditional accuracy and high joint consistency without learned tokenizers. We then apply the same protocol at larger scale to speech--text and protein sequence--structure modeling. Trained without text pretraining, a single speech--text model performs speech synthesis, recognition, continuation, and unconditional co-generation. Jointly generated speech and text achieve 3.5\% word error rate between the generated text and automatic transcriptions of the generated speech. In protein modeling, a single CoBit checkpoint supports forward folding, inverse folding, and joint sequence--structure generation. For jointly generated proteins spanning 100--500 residues, computational refolding of CoBit's sequences yields structures that agree with the generated backbones at a mean self-consistency TM-score of $0.911$. Together, these results show that CoBit provides a common generative framework for substantially different modalities and forms of cross-modal dependence, without introducing a domain-specific objective or sampling procedure.
\begingroup
\makeatletter
\long\def\@makefntext#1{\noindent\@makefnmark\,#1}
\makeatother
\renewcommand{\thefootnote}{\fnsymbol{footnote}}
\footnote[3]{Code available at
\url{https://github.com/GBATZOLIS/BitstreamDiffusion}}
\endgroup
\end{abstract}
\section{Introduction}
\label{sec:intro}

Multimodal generative models must reconcile data with very different structures. Images and audio
are commonly modelled through continuous diffusion, whereas language and protein sequences are
usually generated autoregressively or with discrete corruption processes
\citep{ho2020denoising,song2021score,austin2021structured,sahoo2024simple,wang2024dplm}. Existing
systems bridge these choices through modality-specific encoders, latent spaces, objectives or
prediction interfaces \citep{bao2023unidiffuser,tang2023codi,campbell2024multiflow,wang2025dplm2}.
These designs are effective, but each new modality pairing introduces another interface to design
and calibrate.

In this work, we ask whether one denoising interface and objective can instead model joint distributions over discrete codes from very different modality pairs, with domain-specific choices confined to tokenizers, layouts and inference settings. Bit sequences are the fundamental language of digital computers: a fixed binary code needs no learned embedding and its width grows only logarithmically with the vocabulary, so codes from different vocabularies can share one input projection and one prediction head. Our starting point is therefore continuous bitstream diffusion \citep{method:cobit}, which writes discrete codes as bits and applies continuous Gaussian diffusion to them. Each modality is written as bits, directly or through a frozen, domain-specific tokenizer, placed in a shared binary vocabulary and arranged in a fixed sequence template. Inside the denoiser, every position uses the same input
projection, transformer backbone, bitwise prediction head and loss. A partial-denoising objective
then trains one checkpoint to perform joint generation and every conditional direction included in
training: clean positions specify the conditioning signal, and corrupted positions specify what to
generate. To test whether the denoiser itself needs modality-specific components, the architecture contains no modality- or segment-specific parameters.

We test this question in three settings that differ in the kind of cross-modal dependence they contain. MNIST-Sum pairs raw binary pixels with symbolic equations and uses no learned tokenizer, so it isolates whether visual and symbolic bits form one learnable joint distribution. Speech and text are strongly asymmetric: speech carries acoustic and speaker information absent from the transcript and reaches the model through a lossy codec. Protein sequence and structure are aligned residue by residue and physically coupled rather than semantically related. Every experiment keeps CoBit's objective and its training and sampling protocol, developed for language modeling; our contribution is evidence that this protocol learns joint distributions across these three kinds of dependence.

\section{Related Work}
\label{sec:related}

\textbf{Continuous generation of discrete data.} Beyond discrete corruption \citep{austin2021structured,sahoo2024simple}, continuous methods differ in the state they generate. Diffusion-LM \citep{li2022diffusionlm}, continuous categorical diffusion \citep{dieleman2022continuous}, LangFlow \citep{chen2026langflow} and ELF \citep{hu2026elf} generate learned token embeddings and map them back to tokens with a learned readout; Categorical Flow Maps learn a flow map towards the probability simplex and distil it for few-step categorical generation \citep{roos2026categorical}. Analog Bits writes each token as a fixed binary code and diffuses its bits as real values \citep{chen2023analog}; CoBit develops this into a language model with a binary score-matching objective, an analytic per-bit posterior term and a data-derived noise schedule \citep{method:cobit}. Each targets a single domain; we retain the binary core for multimodal co-generation.

\begin{table}[t]
\centering
\footnotesize
\setlength{\tabcolsep}{3.5pt}
\begin{tabular}{@{}llllcl@{}}
\toprule
Method & Data & Generated state & Objective & Shared I/O & Directions \\
\midrule
UniDiffuser & image--text & continuous latents & noise prediction & \xmark & joint, cond., marginal \\
Show-o & image--text & discrete tokens & AR + masked diffusion & \cmark & cond., mixed-modality \\
Spirit-LM & speech--text & discrete tokens & autoregressive & \cmark & set by prompting \\
DPLM-2 & protein & discrete tokens & discrete diffusion & \cmark & joint, cond., marginal \\
\midrule
CoBit (this work) & all three$^\dagger$ & fixed binary codes & binary score matching & \cmark & joint, cond. \\
\bottomrule
\end{tabular}
\caption{Representative multimodal generators. Shared I/O: one input layer and one output head serve all
modalities of a model. $^\dagger$One model per pair, one shared training and sampling protocol.}
\label{tab:related}
\end{table}

\textbf{Multimodal generation.} UniDiffuser uses one diffusion transformer with modality-specific
times, while CoDi composes modality-specific latent diffusion models
\citep{bao2023unidiffuser,tang2023codi}. MUNI learns a shared stochastic latent with modality-specific encoders and decoders for subset-conditioned and unconditional joint generation \citep{yeo2026muni}. Other systems mix objectives, predict visual features or autoregress over visual tokens for images and text \citep{image:showo,image:seedx,image:lwm}, use textless, text-pretrained or flow-matched speech streams \citep{audio:audiolm,audio:twist,audio:spirit-lm,audio:moshi,audio:taste,audio:flow-slm}, or pair discrete sequence generation with geometric flows, structure tokens or multi-track masking for proteins \citep{campbell2024multiflow,wang2025dplm2,protein:esm3}. Table~\ref{tab:related} compares representative systems. Show-o, Spirit-LM and DPLM-2 already share one embedding table and softmax across modalities, with parameters per token; CoBit shares one input projection and bitwise head over fixed codes and keeps one training and sampling protocol across three domains.

\textbf{Information-based scheduling.} Entropic scheduling allocates steps according to the rate
at which diffusion destroys information, and related estimators select training noise levels
\citep{stancevic2025entropic,raya2026infonoise,trentini2026entropybridgeconditionalmarginaldiscretization}.
CoBit estimates this profile online. We reuse the estimator for each joint bitstream rather than
designing a separate schedule for each modality.

\section{Method}
\label{sec:method}

We extend CoBit \citep{method:cobit} to learn a joint distribution over modalities represented
as bits. The extension has two parts: a layout that brings their representations into one
bitstream, and a training mixture in which the observed bits determine what the model must
generate. One denoiser can then generate paired observations together or complete one from the
other. We retain CoBit's diffusion process, denoiser parameterisation and scheduling rule;
our contribution is their use in a shared multimodal formulation, evaluated across domains.

\subsection{Representing modalities in one bitstream}
\label{sec:method:multimodal}

Each example is encoded as $\vx_0\in\{0,1\}^{S}$. Discrete symbols are written as binary
codes, while continuous observations first pass through a frozen tokenizer. A fixed layout, that is a sequence template chosen per modality pair, sets the order, position and token budget of each modality's segment or aligned channel and the marker and padding positions; longer inputs are truncated, shorter ones padded, and no learned module decides where a code is placed. Tokenizers and decoders are modality-specific; the denoiser operates only on their bits.

The layout can preserve different kinds of correspondence. For speech and text, we concatenate
segments of text, speaker and speech codes. Their vocabularies occupy disjoint integer ranges
and fit into a common 18-bit word. MNIST-Sum instead combines raw binary pixels with deterministic
7-bit text codes, requiring no learned tokenizer. Protein sequence and structure have a natural
residue-level alignment: we concatenate the five amino-acid bits and thirteen structure-code
bits at each residue.

Following CoBit, we group the stream into fixed-width patches and project each patch into one
transformer position. A patch can hold a token, several symbols or aligned codes from two
modalities. All positions share one input projection, transformer trunk and bitwise prediction
head, with one-dimensional positional information. The layout supplies modality identity through
position and code assignment, without a learned modality-specific embedding. Patch widths,
encodings and budgets are specified with the experiments.

\subsection{Continuous denoising with CoBit}
\label{sec:method:cobit}

CoBit treats binary values as continuous during diffusion, applying Gaussian corruption
$\vx_\sigma=\vx_0+\sigma\bm{\epsilon}$, where
$\bm{\epsilon}\sim\mathcal N(\vzero,\mathbf I)$.
The denoiser $D_\theta(\vx_\sigma,\sigma)\in(0,1)^S$ estimates the probability that each
clean bit is one. Its logits combine an analytic isolated-bit posterior with a learned residual
that captures dependencies across the stream. Here those dependencies include both the structure
within each modality and the correspondence between modalities. Self-conditioning supplies a
previous clean estimate as an additional input; we suppress this argument in the notation.

We also reuse CoBit's online entropy-derived schedule to allocate training noise levels and
sampling steps according to the measured denoising error. All bits being generated share a
noise level, and the schedule is estimated for each checkpoint's training mixture.
Appendix~\ref{app:method} gives the parameterisation and scheduling details.

\subsection{Joint and conditional generation by partial denoising}
\label{sec:method:directions}

For each training example, we sample a task and construct a mask $M\in\{0,1\}^{S}$, with
$M_i=1$ for observed bits and $M_i=0$ for bits to generate. Observed bits remain clean:
\begin{equation}
\vx_\sigma=\vx_0+(1-M)\odot\sigma\bm{\epsilon}.
\label{eq:partial-denoising}
\end{equation}
We apply CoBit's binary score-matching objective only to the generated bits,
\begin{equation}
\Ls(\theta)\propto
\E\!\left[w(\sigma)
\left\|(1-M)\odot\bigl(D_\theta(\vx_\sigma,\sigma)-\vx_0\bigr)\right\|^2\right],
\qquad
w(\sigma)=\frac{\sigma^2+\sigma_{\mathrm{data}}^2}
{\sigma^2\sigma_{\mathrm{data}}^2},
\label{eq:loss}
\end{equation}
where $\sigma_{\mathrm{data}}=1/2$ and the expectation includes examples, tasks, noise levels
and Gaussian noise. The weighting follows EDM \citep{image:edm}; no separate loss is introduced for either modality.

Leaving all content bits unobserved trains joint generation. Observing one modality trains
generation of the other, while observing a prefix trains continuation. The mask acts on bits,
so it can select an entire segment or one channel within a protein patch. Fixed structural
markers, where used, remain observed even for joint generation. Each checkpoint is trained on
the task mixture specified for its experiment.

At inference, we initialise the unobserved bits with Gaussian noise and hold the observed bits
fixed throughout denoising. A first-order probability-flow solver updates all generated bits in
parallel, optionally with stochastic churn. Conditional sampling can use classifier-free guidance
\citep{method:cfg}, combining predictions with the observed content present and replaced by a
null value of $1/2$. Guided steps use two denoiser evaluations; unguided steps use one.
Final predictions are converted back to codes and decoded with the corresponding modality
decoder.

\section{Experiments}
\label{sec:exp}

\subsection{Experimental setup}
\label{sec:exp:setup}

We train a separate denoiser from scratch for each modality pair and use the same checkpoint for all
directions reported for that pair. All experiments use CoBit's training and sampling protocol (\S\ref{sec:method}):
Gaussian corruption of the bitstream, the binary score-matching objective of \eqref{eq:loss}, an
online entropy-derived schedule, self-conditioning and a first-order sampler
on the resulting grid. They differ only in their frozen tokenizers and decoders, sequence layouts, task mixtures,
backbone sizes and training hyperparameters, training data and length, and sampler settings. We report both integration steps and
network function evaluations (NFE). Evaluation instruments are domain-specific and are described with each experiment.

\subsection{Image and text: a controlled proof of concept}
\label{sec:exp:imagetext}

\begin{figure}[t]
\centering
\includegraphics[width=\linewidth]{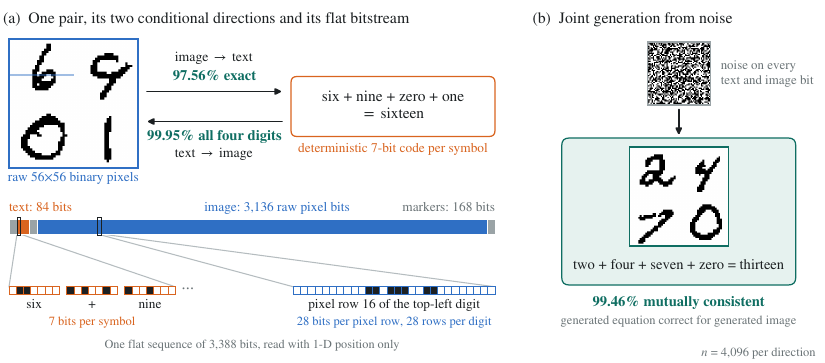}
\caption{MNIST-Sum, a controlled image--text evaluation. (a) Four binarised MNIST digits and their equation form one flat sequence of 3,388 bits (3,136 pixel, 84 text and 168 marker bits); the zoom-ins show bits from the displayed pair. One denoiser serves both conditional directions by observing one part of the stream and denoising the other. (b) With nothing observed, the same weights co-generate an image and equation from noise; a pair is correct only when the generated equation matches the generated image. Rates use $n{=}4{,}096$ samples per direction; the displayed examples illustrate, but do not estimate, those rates.}
\label{fig:mnist:sum}
\end{figure}

MNIST-Sum is a controlled test of the multimodal interface, not a natural-image benchmark. Each
example pairs four binarised MNIST digits \citep{image:mnist} with their ordered equation and sum.
The model must read and add a supplied image, draw the four digits named by a supplied equation, or
generate an image and equation that agree. Neither modality uses a learned tokenizer, so there is
no reconstruction floor. Image$\rightarrow$text is scored by exact string match; image-emitting
directions are evaluated with a fixed, separately calibrated digit classifier. The task therefore
provides an unusually direct test of whether raw visual bits and symbolic text bits can share one
generative space under a modality-agnostic 1-D denoiser.

\textbf{One bitstream, no learned tokenizer.} Images enter as raw binary pixels and text uses a
deterministic 40-symbol, 7-bit code. Each pair becomes a sequence of 3,388 bits: 3,136 pixel bits,
84 text bits and 168 marker bits, grouped into 28-bit positions. The denoiser receives this
sequence and its 1-D positions, but no pixel grid, 2-D attention pattern, modality-specific
embedding, encoder or prediction head. A single 63.0M-parameter model, trained for 500K steps,
performs all three directions; only the set of observed positions changes.

\textbf{Result.} At $n{=}4{,}096$ per direction, image$\rightarrow$text is 97.56\% exact,
text$\rightarrow$image places all four requested digits correctly in 99.95\% of samples, and
joint co-generation from noise is 99.46\% mutually consistent (Figure~\ref{fig:mnist:sum}).
Held-out digit tuples reach 97.85\% image$\rightarrow$text, statistically indistinguishable from
seen tuples. The residual joint error splits roughly evenly between cross-modal disagreement and arithmetic
slips, each under 0.3\%. Figure~\ref{fig:mnist:sum} illustrates the outputs, while
Appendix~\ref{app:image:mnist} reports classifier calibration, confidence intervals and the error
decomposition. With CoBit's language-modeling protocol, raw visual and symbolic bits thus form one learnable joint distribution; learned tokenizers and open-ended outputs enter with the speech--text (\S\ref{sec:exp:audio}) and protein (\S\ref{sec:exp:protein}) experiments.

\FloatBarrier

\subsection{Audio and text}
\label{sec:exp:audio}

Joint text-audio modeling is precisely the kind of discrete-continuous mismatch problem for which bitstream diffusion is well suited: text represents an utterance as a discrete sequence and speech as a continuous signal with richer information, including speaker identity, emotion and background noise. The relationship between the two modalities can be easily measured through the alignment of their content. We follow the speech language modeling paradigm \citep{audio:twist, audio:spirit-lm}, where speech is tokenized into discrete units and generation is performed over text and speech tokens. We demonstrate how CoBit works as a non-autoregressive speech language model (SLM) that supports joint generation, text-to-speech synthesis, automatic speech recognition and speech continuation with a single set of weights.

\textbf{Setup.} We train two models: CoBit-LibriTTS (134M, $12\times768$) on LibriTTS \citep{audio:libritts} (354K pairs, 140K steps) and CoBit-MLS (633M, $26\times1152$) on both LibriTTS and MLS-English \citep{audio:mls} (11.2M pairs, $1.2\times10^6$ steps). We quantize audio with the FSQ-based StableCodec \texttt{1x46656}, which produces 25 tokens per second (budget of 800 and 500 tokens, respectively), combined with a fixed-length (32) global speaker representation from BiCodec-global \citep{audio:spark-tts}. Speaker tokens are included as a conditioning signal for generation and are omitted during audio decoding. Using \texttt{o200k} for text (budget of 168 and 100 tokens, respectively), the full vocabulary fits into 18 bits. Training additionally includes a prompt-conditional regime, with equal effort (0.25) allocated to each setting. Evaluation is conducted on LibriSpeech-PC \citep{audio:librispeech-pc} and the SALMON benchmark. Text pre-training and speaker conditioning on the decoder are omitted. Warm-initializing CoBit from a pre-trained text checkpoint is straightforward in principle: it would mainly require adapting the components that are tied to the sequence length and the vocabulary bit-width. We expect both changes to close the remaining performance gap and leave their implementation to a dedicated line of work on text-audio CoBit.

\begin{figure}[ht]
    \centering
    \includegraphics[width=\linewidth]{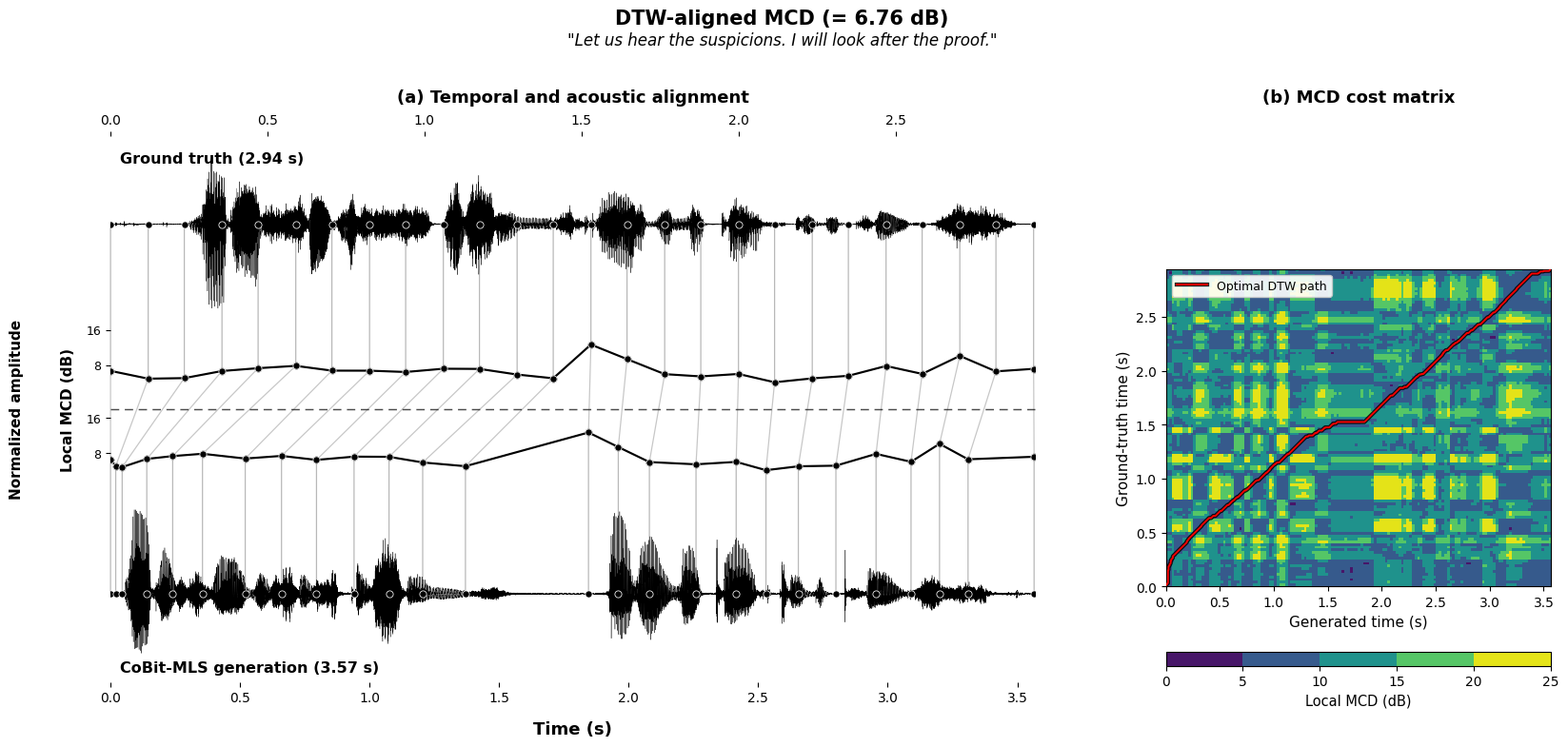}
    \caption{Text-to-speech generation with CoBit-MLS. Given text and speaker tokens (extracted from a held-out example), the model is tasked with generating the speech part of the sequence. The similarity between the reference and generated waveforms is measured in terms of frame-level mel-cepstral distortion (MCD), aligned using dynamic time warping (DTW). CoBit generates the same text, but with different speech pacing and a longer pause between the two sentences. (Left) Temporal alignment of waveform frames. CoBit introduces silence between the two sentences. (Right) Pairwise MCD cost matrix and the optimal DTW path. The horizontal segment between 1.5 and 2 seconds corresponds to inter-sentence silence.}
    \label{fig:audio:tts-dtw}
\end{figure}

\textbf{Results.} Co-generation and conditional directions are reported in Table \ref{tab:audio:headline}, using the same weights for all tasks. CoBit-MLS can co-generate text and audio consistently, achieving a cross-modal WER of 3.5. A UTMOS score of 3.77 suggests natural-sounding audio, despite sampling the speaker identity from scratch. The content also remains highly plausible and syntactically solid in terms of perplexity under an external LM. Both models are considerably stronger in speech-emitting than text-emitting generation. The similarity between ground truth and generated text-to-speech samples is visualized in Figure \ref{fig:audio:tts-dtw}. In speech continuation, CoBit samples acoustically and linguistically grounded audio, yet it struggles to preserve thematic content. StableCodec sets an upper bound for audio performance, with CoBit-MLS recovering almost reconstruction-level SpkSim for speech continuation.\footnote{Inspect text-audio-examples online: \url{https://demo-2546.github.io/cobit-text-audio-samples/}.}

\begin{table}[t]
\centering
\footnotesize
\setlength{\tabcolsep}{4pt}
\resizebox{\linewidth}{!}{
\begin{tabular}{lrrrrrrrr}
\toprule
\textbf{Direction} & $n$ & Steps (NFE) & \textbf{WER}$\downarrow$ & \textbf{CER}$\downarrow$ & \textbf{UTMOS}$\uparrow$ & \textbf{SpkSim}$\uparrow$ & \textbf{FSD}$\downarrow$ &\textbf{GenPPL}$\downarrow$ \\
\midrule
\multicolumn{7}{l}{\emph{CoBit-LibriTTS --- one set of weights (134M, 800 speech tokens)}}\\
text $\rightarrow$ speech & 2{,}294 & 512 (1024) & 11.7 & 8.1 & 3.31 & 0.19 & --- & --- \\
speech $\rightarrow$ text & 2{,}294 / 2{,}813 & 512 (1024) & 23.3 / 38 & 15.3 / 24.1 & --- & --- & --- & --- \\
speech continuation & 1{,}936 & 512 (1024) & --- & --- & 2.82 & 0.21 & 2,373 / 40.5 & 152.1 \\
joint & 1{,}024 & 256 (256) & 19.9 & 13.3 & 3.02 & --- & --- & 509.6 / 377.2 \\
\midrule
\multicolumn{7}{l}{\emph{CoBit-MLS --- one set of weights (633M, 500 speech tokens)}}\\
text $\rightarrow$ speech & 2{,}294 & 512 (1024) & 5.3 & 4.1 & 4 & 0.27 & --- & --- \\
speech $\rightarrow$ text & 2{,}294 / 2{,}813 & 512 (1024) & 8.2 / 22 & 4.3 / 13.1 & --- & --- & --- & --- \\
speech continuation & 1{,}936 & 512 (1024) & --- & --- & 3.89 & 0.34 & 2,386 / 13 & 83.6 \\
joint & 1{,}024 & 512 (512) & 3.5 & 1.8 & 3.77 & --- & --- & 252.4 / 133.3 \\
\bottomrule
\end{tabular}
}
\caption{All text-audio generative directions from a single set of weights per model. Each conditional row is quoted at the best step count for that direction, evaluated on LibriSpeech-PC. Generated audio is transcribed with Whisper-large. Per-direction guidance and churn are listed in Appendix \ref{app:audio:protocol}; co-generation uses $\gamma=0.175$. Text-to-speech and speech continuation evaluation is on test-clean; the two scores for speech-to-text refer to test-clean and test-other, respectively. For FSD, the two separate metrics refer to FSD-wlm and FSD-e2v, respectively. For co-generation, generative perplexity is reported for both the text (left) and the audio (right) streams.}
\label{tab:audio:headline}
\end{table}

\textbf{Comparison to prior work.} Appendix \ref{app:audio:results} places CoBit among autoregressive SLMs and other baselines for all tasks. The capabilities of bitstream diffusion are best demonstrated in speech continuation, where CoBit-MLS outperforms all baselines except TASLM-Embed across audio-quality based metrics and generative perplexity. Modifying the sampler settings, CoBit-MLS can surpass even TASLM-Embed in terms of expressiveness (Appendix \ref{app:audio:sampler}).  LLM-as-a-Judge is CoBit's weakest overall evaluation (see Appendix \ref{app:audio:llm} for further discussion). In SALMON-based evaluation, CoBit-MLS scores better on average than all other text-speech SLMs, but cannot match speech-only models. While it remains competitive on the voice-related tasks, it is worse than Mimi-based models such as Flow-SLM and Llama-Mimi on noise-related ones. In text-to-speech, CoBit-MLS surpasses SpiritLM on CER (4.1 versus 6.7) and approaches Moshi on WER (5.3 versus 4.7), despite being 10 times smaller. However, despite the consistency and naturalness of generated audio, speaker characteristics are not reproduced to the same level as by the ASR-TTS hybrid GPA-v1.5 \citep{audio:gpa} where the decoder is conditioned on the speaker. Speech recognition is the task where the absence of text-only training matters the most, with CoBit having the worst performance relative to baselines.

\textbf{Sampler: integration steps and guidance aid semantics.} Additional NFE are most critical when the model is tasked with producing novel semantics (joint generation and speech continuation): generative perplexity decreases near-monotonically as the step count increases. For cross-modal tasks, on the other hand, performance plateaus at 32 integration steps, with higher counts offering mild improvements. The same perplexity pattern for speech continuation is observed when increasing the guidance scale as well. Guidance also plays a crucial role for alignment in cross-modal directions, especially text-to-speech. However, increasing the guidance scale has a negative effect on frame-level fidelity and expressiveness: minimal and no guidance settings score the lowest reported FSD-e2v figure surpassing all baselines, along with improved FSD-wlm scores (Appendix \ref{app:audio:sampler}).

\textbf{Limitations.} Thematic extrapolation is the model's worst aspect, with CoBit continuations often receiving the lowest possible score. Speech recognition is the overall weakest task, with a large portion of the errors occurring at word-level (CoBit-MLS: WER $\simeq2\times$ CER on test-clean). These limitations are strongly tied to the lack of text pre-training. Imprecise speaker replication is directly associated with the low SpkSim ceiling of StableCodec reconstructions and, for text-to-speech in particular, the lack of speaker conditioning on the audio decoder. While such design choices are well-established in the SLM literature, they are not included in this work to demonstrate what the generalized bitstream diffusion framework can achieve off-the-shelf.

\begingroup
\raggedbottom

\subsection{Protein sequence and structure}
\label{sec:exp:protein}

Unlike text, image, and audio multi-modality, whose correspondence is primarily semantic, a protein sequence and its three-dimensional structure are physically coupled descriptions. We ask whether the bitstream formulation can learn the coupled modalities of protein sequence and structure by training CoBit from scratch on 216,237 PDB/AFDB-SwissProt \citep{protein:pdb, protein:afdb} pairs with 20\% sequence replay. Each residue aligns five amino-acid bits with thirteen signs from the frozen DPLM-2 lookup-free quantiser (LFQ) \citep{protein:dplm2}. Decoding retains the 20 canonical amino acids and all $2^{13}$ structural codes. The common binary interface lets one denoiser learn sequence--structure dependencies while the frozen codec recovers backbone coordinates. Data, splits, and tokenisation are detailed in Appendices~\ref{app:protein:protocol} and~\ref{app:protein:training}.

\begin{figure}[!t]
    \centering
    \includegraphics[width=.8\linewidth]{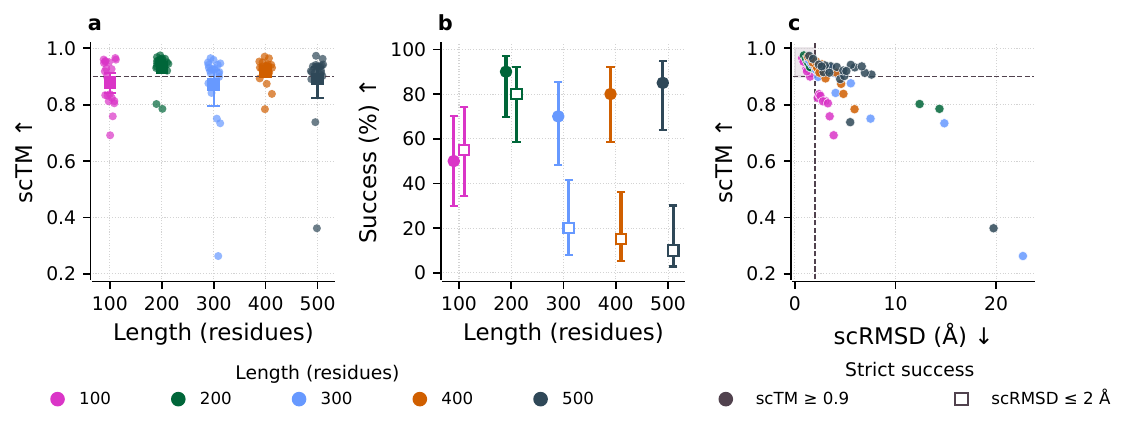}
    \caption{Backbone quality from denoised bits: (a) scTM by length, (b) strict success
    by length, and (c) fold similarity versus coordinate error. Colours denote length. 100 backbones (20 per length) each receive one ProteinMPNN redesign and ESMFold refold. Intervals are 95\%. Shading marks both strict criteria.}
    \label{fig:protein:multimodal}
    \label{fig:protein:strict}
\end{figure}

\textbf{Coverage.} We test bidirectional learning through one shared bitstream objective. Clamping either channel gives inverse or forward folding while leaving both free enables joint generation, without task-specific fine-tuning or multiple sequence alignments as input (Appendix~\ref{app:protein:metrics}). Our evaluations were based on entropic DDIM with $50$ grid points, conditional CFG $5$ and no unconditional guidance. Under these settings, mean recovery is $44.10\%$ on 128 CATH 4.3 validation targets and forward TM is $.554$ on $183$ CAMEO targets \citep{protein:cath,protein:cameo,protein:tmscore}. For $100$ generated backbones ($20$ per length, 100--500), one ProteinMPNN redesign and ESMFold refold give mean scTM $.899$ \citep{protein:proteinmpnn,protein:esmfold}. Figure~\ref{fig:protein:multimodal} shows fold similarity across lengths while Figure~\ref{fig:protein:bestgallery} shows the best backbones per length-bucket. We assess backbone recovery after redesign and refolding using scTM  $\geq.9$ ($D_{\mathrm{TM}}$) for overall fold agreement and scRMSD $\leq2$\,\AA{} ($D_{\mathrm{RMSD}}$) for close coordinate agreement. Full metrics and sampling analysis are in Appendices~\ref{app:protein:metrics} and~\ref{app:protein:pareto}. 

Directly refolding CoBit's own sequences for the same 100 pairs gives mean scTM $.911$, with $D_{\rm TM}=69\%$ and $D_{\rm RMSD}=41\%$ (Table~\ref{tab:protein:direct}; Appendix~\ref{app:protein:direct}). This supports sequence--structure consistency at the fold level, despite lower coordinate-level success.

\par\addvspace{6pt}\noindent
    \begin{minipage}{\linewidth}
    \makeatletter\def\@captype{figure}\makeatother
    \centering
    \includegraphics[width=.99\linewidth]{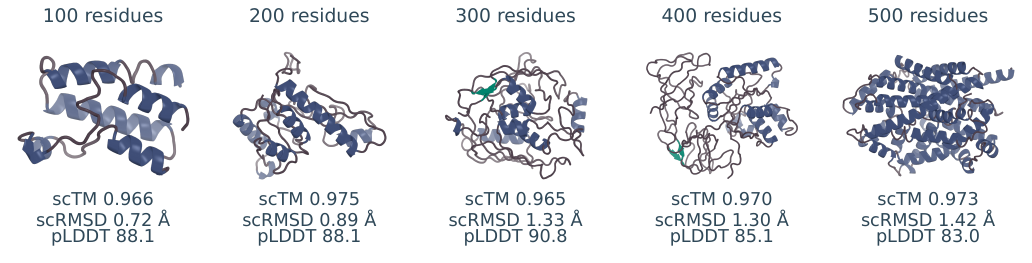}
    \caption{Selected CoBit backbones, rendered with PyMOL \citep{protein:pymol}.
    Highest scTM among 20 samples at each displayed length with scRMSD $\leq2$\,\AA{};
    scores use ProteinMPNN/ESMFold. pLDDT is refold confidence.}
    \label{fig:protein:bestgallery}
\end{minipage}\par\addvspace{6pt}

\textbf{Bitstream capability versus protein optimisation.} We compare CoBit's bitstream formulation against protein-specific models. Table~\ref{tab:protein:baselines} places CoBit alongside the models evaluated by \citet{protein:simpledesign}, following their published one-redesign procedure with 100 samples. MultiFlow \citep{campbell2024multiflow} leads on the strict criteria with a 21.8M-parameter model that uses explicit geometric learning and designability distillation. On the other hand, CoBit tests the common bit objective with an existing codec and paired-data recipe. Its sequence replay supplies unpaired examples unlike MultiFlow's designability distillation and without direct geometric supervision. Appendix~\ref{app:protein:comparison} explains the training differences and MultiFlow's distillation ablation in detail.

\par\addvspace{6pt}\noindent
\begin{minipage}{\linewidth}
\makeatletter\def\@captype{table}\makeatother
\centering
\begin{minipage}[t]{.55\linewidth}
\vspace{0pt}
\caption{Backbone designability (\%) with one ProteinMPNN redesign and ESMFold.
Training data and evaluation settings differ across studies. Published PMPNN1 baselines: \citet{protein:simpledesign}; G: general bitstream framework; P: protein-specific generator.}
\label{tab:protein:baselines}
\label{tab:protein:literaturemain}
\centering
\fontsize{8}{9.5}\selectfont
\setlength{\tabcolsep}{2pt}
\begin{tabular}{@{}rlcrrr@{}}
\toprule
ID & Model & Scope & $n$ & $D_{\rm RMSD}$ & $D_{\rm TM}$ \\
\midrule
\multicolumn{6}{@{}l}{\emph{Our evaluation: joint bitstream generator}} \\
1 & CoBit & G & 100 & 36.0 & 75.0 \\
\midrule
\multicolumn{6}{@{}l}{\emph{Published evaluation: Lu et al. (2026)}} \\
\addlinespace
\multicolumn{6}{@{}l}{Joint generators: discrete structural codes} \\
2 & DPLM-2 (650M) & P & 100 & 31.0 & 48.0 \\
3 & ESM3 Open (sequence first) & P & 100 & 17.0 & 19.0 \\
4 & ESM3 Open (structure first) & P & 100 & 3.0 & 4.0 \\
\addlinespace
\multicolumn{6}{@{}l}{Joint generators: explicit / partially latent geometry} \\
5 & MultiFlow & P & 100 & 86.0 & 90.0 \\
6 & La-Proteina (no-triangle) & P & 100 & 84.0 & 86.0 \\
7 & La-Proteina (triangle) & P & 100 & 84.0 & 88.0 \\
8 & SimpleDesign & P & 100 & 44.0 & 63.0 \\
9 & ProtPardelle & P & 100 & 42.0 & 41.0 \\
10 & ProtPardelle-1c & P & 100 & 52.0 & 53.0 \\
11 & ProteinGenerator & P & 100 & 42.0 & 46.0 \\
\addlinespace
\multicolumn{6}{@{}l}{Structure generators} \\
12 & Proteina & P & 100 & 46.0 & 50.0 \\
13 & RFdiffusion & P & 100 & 49.0 & 54.0 \\
14 & FrameFlow & P & 100 & 46.0 & 49.0 \\
\bottomrule
\end{tabular}
\end{minipage}\hfill
\begin{minipage}[t]{.43\linewidth}
\vspace{0pt}
\makeatletter\def\@captype{figure}\makeatother
\caption{Visual representation of the rates from Table~\ref{tab:protein:baselines}. CoBit captures how proteins fold overall, but is less accurate on fine structural detail.}
\label{fig:protein:modelmap}
\centering
\includegraphics[width=\linewidth]{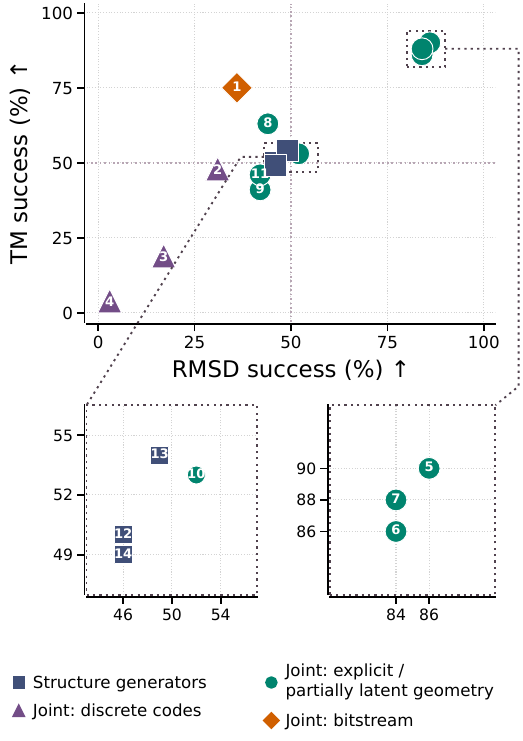}
\end{minipage}
\par\smallskip
\end{minipage}\par\addvspace{6pt}

\textbf{Modeling quality versus biological function.} A dedicated protein study can further test scalability to biological tasks and the effectiveness of physical coupling, for example in binder--linker design with laboratory (wet-lab) validation. CoBit's stronger fold-level ($D_{\mathrm{TM}}$) than coordinate-level ($D_{\mathrm{RMSD}}$) success indicates that denoising structural bits captures useful protein organization information and that geometric precision could be further improved through dedicated protein training, including architectural scaling, hyperparameter tuning, distillation, and additional data. In this work, coordinate-level success varies with protein length as shown in figure~\ref{fig:protein:strict} and, although the present metrics indicate modeling quality, establishing biological function is beyond the scope of this work. 

\endgroup

\section{Limitations}
\label{sec:limitations}

The experiments explore a limited part of the approach's potential. We train the denoisers from
scratch, without the text pretraining used by several competing speech--text systems. This may
particularly affect speech recognition and the semantic coherence of continuations, where paired
speech data must supply linguistic knowledge as well as cross-modal alignment. Text-pretrained
initialisation is therefore a promising direction, although controlled comparisons are needed to
establish its contribution. Protein-sequence pretraining may offer analogous benefits for
sequence--structure modelling.

Our models are also generally smaller than the strongest systems for the corresponding tasks,
and the training datasets provide limited coverage of the distributions we aim to model. For
example, the speech experiments rely on read speech, offering relatively narrow linguistic and
acoustic diversity. These choices restrict what the present results can establish about
performance at scale. Larger backbones and broader, higher-quality datasets are plausible routes
to improvement, but their effects need to be measured within this formulation.

Continuous modalities introduce a further dependence on pretrained tokenizers and decoders.
Information discarded during tokenisation is unavailable to the denoiser, while reconstruction
errors can persist even when codes are modelled accurately. The observed performance therefore
reflects both the bitstream model and its representation. Codec reconstruction results help
distinguish these effects, but do not by themselves explain the remaining generation errors.

\section{Discussion and conclusion}
\label{sec:discussion}
\label{sec:conclusion}

CoBit's training and sampling protocol, developed for language modeling, carries over to three
distinct forms of cross-modal dependence without a new objective or sampling procedure. On
MNIST-Sum it learns visual--symbolic dependence directly from raw bits, with near-perfect
conditional accuracy and 99.46\% mutually consistent joint samples. On speech and text it handles
an asymmetric, codec-backed pair, co-generating speech and text at 3.5\% cross-modal WER, while
speech recognition and thematic continuation remain its weakest directions. On protein sequence and
structure it captures residue-aligned, physically coupled channels: 69\% of its generated pairs
agree at the fold level (scTM~$\geq$~0.9), against 41\% at scRMSD~$\leq$~2\,\AA. Within each
setting, one checkpoint serves joint and conditional generation.

These results come from denoisers of 63M to 633M parameters trained from scratch. We leave to future work how far pretraining, larger models, broader data and better continuous-modality tokenizers can improve performance within the same protocol.

\bibliography{iclr2027_conference}
\bibliographystyle{iclr2027_conference}

\appendix
\section{Additional method details}
\label{app:method}

This appendix records implementation details inherited from text-only CoBit
\citep{method:cobit}. The main paper focuses on the multimodal representation and partial-denoising
objective used in this work.

\subsection{Continuous bitstream diffusion}

CoBit uses a variance-exploding Gaussian diffusion with, by default,
$\sigma\in[0.002,80]$ (CoBit-MLS trains with $\sigma_\text{min}=0.01$), data centre $1/2$ and $\sigma_{\mathrm{data}}=1/2$. The analytic posterior
logit of an isolated bit under a uniform prior,
$(\vx_\sigma-\tfrac12)/\sigma^2$, is clipped to $\pm30$ and added to a learned contextual
residual. The network therefore concentrates on dependencies between bits rather than relearning
the local binary denoising problem.

The entropy-derived schedule estimates the conditional entropy rate from the unweighted
per-example error, $\hat h(\sigma)\propto e(\sigma)/\sigma^2$. Estimates are stored in a FIFO buffer
of $8\times10^5$ examples and 128 log-spaced bins. The training proposal and sampling grid use
$q(\sigma)\propto g(\sigma;c,n)\hat h(\sigma)^{1/2}$, where
$g(\sigma;c,n)=\sigma^n/(\sigma^n+c^n)$ with $c=0.1$ and $n=3$. Training begins with the EDM
log-normal proposal ($P_{\mathrm{mean}}=-1.2$, $P_{\mathrm{std}}=1.2$), interpolates to $q$ over
steps $4$--$5\times10^4$ by default (CoBit-LibriTTS warms up over $1$--$1.3\times10^4$ steps), and refreshes the estimate every $2\times10^3$ steps.

For self-conditioning, the denoiser receives a second input $\hat\vx_0$. With probability
$p_{\mathrm{sc}}=0.5$, a detached no-gradient pass supplies this estimate during training;
otherwise it is zero. At inference, \texttt{carry} mode passes the previous prediction to the next
step without another evaluation; the \texttt{refined} mode, used for the protein evaluations, recomputes the estimate
at the new state and doubles the cost per step.

Sampling integrates the probability-flow ODE with a first-order DDIM/Euler step on the
entropy-derived grid, optionally with EDM-style churn \citep{method:ddim,image:edm}. Final
probabilities are thresholded at $1/2$ and regrouped into codes. The Sequence Diffusion Transformer
patches the $m$ bits at each position into one trunk token, uses RoPE, SwiGLU and adaLN-zero noise
conditioning, and combines the trunk output with the raw noisy bits in a light skip-connection
head.

\subsection{Representation and partial denoising}

Each frozen tokenizer maps its modality to integer codes and decodes predicted codes after
sampling. Per-modality vocabularies occupy disjoint ranges of one integer code space together with
padding and structural markers. The bit width $m$ covers their union, so namespace information is
part of the same binary word as token content. Markers remain clean throughout training and
sampling. Fixed token budgets are truncated or padded, and two-dimensional grids are flattened in
the order specified by the experiment.

Training draws a conditioning regime for each example. The mask $M$ marks clean conditioning bits
and the loss is evaluated only on $1-M$. When self-conditioning is active, clean conditioning bits
are also copied into $\hat\vx_0$.

At inference, the task fixes $M$. Conditioned positions are clamped before every denoiser call and
their ODE velocity is set to zero. Guidance combines conditional and unconditional predictions in
probability space as $D_g=D_u+w(D_c-D_u)$. The two branches are evaluated as one batch and carry
separate self-conditioning estimates. Joint generation uses one evaluation per step; guided
conditional generation uses two.

\section{Image and text: MNIST-Sum details}
\label{app:image}

\subsection{Protocol}
\label{app:image:mnist}

\textbf{Corpus and representation.} Each example contains four native $28{\times}28$ MNIST digits
in raster order and the corresponding equation, including its sum. We sample $10^6$ examples and
hold out 10\% of the $10^4$ possible digit tuples. Pixels are binarised and flattened digit-major,
then row-major within each digit, so every 28-bit image position is one complete digit row. Text
belongs to a closed 40-symbol language (number words 0--36, ``$+$'', ``$=$'' and padding)
represented by deterministic 7-bit codes, four per 28-bit position. The sequence has 121
positions of 28 bits (3 text, 112 image and 6 markers; 3,388 bits in total). Neither modality uses a learned tokenizer.

\textbf{Training and evaluation.} The $12{\times}512$ denoiser (63.0M parameters) trains for 500K
steps with the shared diffusion objective, entropy controller and self-conditioning, on a task
mixture of 35\% text$\rightarrow$image, 35\% image$\rightarrow$text and 30\% joint generation, and receives
only the flattened sequence and its 1-D position. The positional settings
are those of the speech--text models, with four local Fourier features and an intra-patch period
of $P-1$; there is no segment embedding.
Evaluation uses EMA weights and $n{=}4{,}096$ independently sampled examples per direction and
split. All directions use 256 integration steps of the entropy-derived DDIM sampler with carry
self-conditioning. Image$\rightarrow$text and text$\rightarrow$image use guidance scale 2, so each step evaluates
the denoiser twice (512 NFE); joint generation is unguided. Stochastic churn is $\gamma{=}0.175$ for
image$\rightarrow$text and joint generation and 0 for text$\rightarrow$image. Confidence intervals are 95\% Wilson
intervals.

\subsection{Results, calibration and error decomposition}

\begin{table}[h]
\centering
\small
\setlength{\tabcolsep}{6pt}
\begin{tabular}{lrrr}
\toprule
& Seen tuples & Held-out tuples & CI (seen) \\
\midrule
Image$\rightarrow$text, exact string match & 97.56 & 97.85 & $[97.04, 97.99]$ \\
Text$\rightarrow$image, all four digits & 99.95 & 99.93 & $[99.82, 99.99]$ \\
Joint from noise, mutually consistent & 99.46 & --- & $[99.19, 99.65]$ \\
\bottomrule
\end{tabular}
\caption{MNIST-Sum task accuracy (\%), $n{=}4{,}096$ per direction and split. ``Held-out'' uses digit tuples
absent from training. Joint generation is counted correct only when the generated equation is
arithmetically valid and names the four digits in the generated image.}
\label{tab:app:mnist:complete}
\end{table}

\textbf{Image$\rightarrow$text} is scored by exact string equality, without a learned instrument.
The held-out interval $[97.36, 98.25]$ overlaps the seen interval almost entirely, so the
generalisation gap is not statistically distinguishable from zero. The seen result decomposes
into 97.66\% correct perception of all four digits, 99.90\% correct arithmetic given correct
perception and no malformed output. Per slot, perception is 99.41\%, 99.12\%, 99.58\% and
99.49\% for the top-left, top-right, bottom-left and bottom-right digits. An image-blind model
that always answers the most common sum scores 6.52\% on seen tuples and 7.80\% on held-out
tuples.

\textbf{Text$\rightarrow$image} and the image half of joint generation are read by one fixed
quadrant classifier, which reads the requested digit in 99.99\% of generated quadrants. As a
reference for the instrument itself, it identifies all four digits in 96.09\% (validation) and
96.80\% (held-out) of real composites built from the MNIST test pool, from which both splits are
drawn, and in 99.78\% of composites built from the train pool it was fitted on. These figures are
not upper bounds on the generation scores: generated digits imitate the train pool and are often
more prototypical than real handwriting, so a generation score can exceed them, as
text$\rightarrow$image does (99.95\%). We report them as calibrations and do not use them to
normalise the generation score.

\textbf{Joint co-generation} samples image and text from noise in one trajectory. Of the 4,096
pairs, 100.00\% have a well-formed equation, 99.71\% name the four digits that the classifier
reads in the generated image and 99.76\% are arithmetically valid; the 99.46\% that satisfy all
three is the reported rate. Cross-modal disagreement (0.29\%) and arithmetic error (0.24\%) are
of comparable size, so the residual failures cannot be attributed mainly to either.

\section{Audio and text experiments}
\label{app:audio}

\subsection{Protocol and Checkpoints} \label{app:audio:protocol}

\textbf{LibriVox-based corpora.} LibriTTS, MLS-English, LibriSpeech-PC and part of SALMON (background and room partitions) are extracted from the LibriVox project, and thus consist of single-speaker, reading-style audiobook recordings. The content CoBit sees during training is exclusively story passages, which severely constrains its stylistic and thematic exposure. Availability of in-the-wild text-audio pairs is very limited at this scale for single-speaker samples, so audiobook data is the natural alternative. For most speech models, semantic and stylistic generality is inherited from the text pre-trained backbone. Note that the training and test sets have no example or speaker overlap. 

\textbf{Linguistic evaluation.} To assess the linguistic properties of generated speech signals, we transcribe them using an external Whisper-large \citep{audio:whisper} ASR model. For co-generation, error rates are computed between the two generated sequences. The fluency of a text sequence is measured by its perplexity under an external language model (in this case Llama-3.2 1B), computed separately for each modality in co-generation and for speech transcriptions in speech continuation. To further determine the conceptual relevance of produced continuations, \texttt{GPT-4o} is employed in an LLM-as-a-Judge setting, assigning each continuation a score from 1 to 5. The system message is taken directly from \cite{audio:taste} and presented in Table \ref{tab:audio:llm-judge}.

\begin{table}[t]
\centering
\small
\setlength{\tabcolsep}{4.5pt}
\begin{tabular}{l}
\toprule
The task is evaluating the relevance and likelihood of the predicted text continuation, given the text prompt. \\ 
You should also consider whether the meaning of the text continuation is making sense. \\
The text prompt is: "\texttt{prompt}", and the text continuation is: "\texttt{content}" \\ \\

You must give an overall rating from 1 to 5. The rating guideline is as below: \\
1: The text continuation is very unlikely and irrelevant to the text prompt. \\
2: The text continuation is unlikely and marginally relevant to the text prompt. \\
3: The text continuation is moderately likely and relevant to the text prompt. \\
4: The text continuation is likely and relevant to the text. \\
5: The text continuation is very likely and highly relevant. \\ \\

You should take the following steps to provide the score: \\
First: briefly analyse the sample with the above definition. \\
Second: MUST follow the output format as: I would rate the score as \_ \\
\bottomrule
\end{tabular}
\caption{LLM-as-a-Judge prompt for thematic relevance evaluation of speech continuation. GPT-4o is used as the judge model.}
\label{tab:audio:llm-judge}
\end{table}

\textbf{Acoustic evaluation.} For TTS, voice replication is captured in terms of speaker similarity (SpkSim), computed as the cosine similarity between the outputs of an ECAPA-TDNN \citep{audio:ecapa-tdnn} speaker verification model. For speech continuation, we are also interested in how closely the generated audio follows the distribution of ground-truth continuations. We thus measure Fr\'echet Speech Distance (FSD), an adaptation of the FID metric to audio. Population statistics are extracted from self-supervised representations of the true and generated continuations, and the Fr\'echet distance between the resulting distributions is calculated. Following previous work \citep{audio:flow-slm}, two variants are considered: FSD-wlm, which is based on embeddings from the 6th layer of WavLM-large \citep{audio:wavlm} to measure frame-level match, and FSD-e2v, which is based on expressiveness representations from Emotion2Vec \citep{audio:emotion2vec}. FSD metrics do not evaluate the quality of the generated audio; they capture its population-level resemblance to the original continuations. FSD metrics are investigated in more detail in Appendix \ref{app:audio:sampler}.

\textbf{Generation-based SALMON benchmark.} The SALMON benchmark is used for acoustic evaluation. In its original form, SALMON creates replicas of short speech recordings where the start remains identical, but some paralinguistic property (sentiment, speaker identity/gender, background, room) is modified midway. SLMs receive an accuracy score based on how often the positive example is assigned higher likelihood than the negative one. We use the generation-based SALMON formulation \citep{audio:gen-salmon} because our implementation does not provide the discrete sequence likelihoods required by the original likelihood based protocol: The examined SLM produces a continuation of the shared start of the two examples, and a judge model calculates embeddings for the positive, negative and generated continuations. The SLM then receives an accuracy score based on how often the generated continuation is closer to the positive example than to the negative one, in terms of cosine similarity in the embedding space. TITANET \citep{audio:titanet} is used as the judge for sentiment, speaker and gender tasks; HuBERT-large-audioset and Wav2Vec2-large-audioset \citep{audio:ARCH} are used for background and room tasks, respectively. 

\textbf{Operating points.} The sampler configurations (integration steps, stochasticity and guidance scale) for each conditional task were selected based on modality alignment (WER) on 500 samples from the LibriTTS dev-clean set; the final settings are summarized in Table \ref{tab:audio:operating-points} and full validation scores are reported in Tables \ref{tab:audio:val:tts}, \ref{tab:audio:val:stt} and \ref{tab:audio:val:cont}.

\begin{table}[t]
\centering
\small
\setlength{\tabcolsep}{4.5pt}
\begin{tabular}{lrrr}
\toprule
\textbf{Direction} & Steps (NFE) & $\gamma$ & gs \\
\midrule
\multicolumn{4}{l}{\emph{CoBit-LibriTTS --- one set of weights (134M, 800 speech tokens)}}\\
text $\rightarrow$ speech & 512 (1024) & 0.23 & 6 \\
speech $\rightarrow$ text & 512 (1024) & 0.37 & 5 \\
speech continuation & 512 (1024) & 0.15 & 5 \\
\midrule
\multicolumn{4}{l}{\emph{CoBit-MLS --- one set of weights (633M, 500 speech tokens)}}\\
text $\rightarrow$ speech & 512 (1024) & 0.27 & 5 \\
speech $\rightarrow$ text & 512 (1024) & 0.27 & 5 \\
speech continuation & 512 (1024) & 0.4 & 5 \\
\bottomrule
\end{tabular}
\caption{Sampler operating points for text-audio experiments. Steps, stochasticity and guidance settings were chosen based on validation performance on 500 samples from LibriTTS dev-clean.}
\label{tab:audio:operating-points}
\end{table}

\begin{table}[t]
\centering
\footnotesize
\setlength{\tabcolsep}{4pt}
\begin{tabular}{l rrrrrr}
\toprule
 & \multicolumn{3}{c}{\textbf{CoBit-LibriTTS}} & \multicolumn{3}{c}{\textbf{CoBit-MLS}} \\
\cmidrule(lr){2-4} \cmidrule(lr){5-7}
$\gamma$ & \textbf{WER}$\downarrow$ & \textbf{CER}$\downarrow$ & \textbf{UTMOS}$\uparrow$ & \textbf{WER}$\downarrow$ & \textbf{CER}$\downarrow$ & \textbf{UTMOS}$\uparrow$ \\
\midrule
\multicolumn{7}{l}{\emph{Steps=256 (NFE=512); gs=5/gs=6}} \\
0.11 & 11.4 / 12.3 & 8.2 / 8.7 & 3.29 / 3.27 & 7.2 / 7.4 & 5.7 / 6.1 & 3.96 / 3.94 \\
0.13 & 12.2 / 12.6 & 8.8 / 8.8 & 3.3 / 3.27 & 6.4 / 7.9 & 5 / 6.4 & 3.95 / 3.96 \\
0.15 & 11.4 / 11.9 & 8.2 / 8.7 & \textbf{3.32} / 3.27 & 6.8 / 7 & 5.5 / 5.4 & 3.97 / 3.95 \\
0.17 & 12.2/ 12 & 8.9 / 8.5 & 3.3 / 3.27 & 5.5 / 6.9 & 4.3 / 5.1 & 3.95 / 3.95 \\
0.19 & 11.6 / 11.3 & 8.2 / 8.2 & 3.3 / 3.27 & 5.4 / 6.6 & 4.2 / 5.2 & 3.96 / 3.94 \\
0.22 & 11.4 / 12 & 8 / 8.5 & 3.27 / 3.25 & 6.5 / 6.6 & 5 / 5.3 & 3.94 / 3.96 \\
0.25 & 12 / 13 & 8.2 / 9.8 & 3.28 / 3.29 & 5.4 / 6.3 & 4 / 4.9 & 3.95 / 3.97 \\
0.28 & 11.9 / 12.2 & 8.2 /8.6 & 3.27 / 3.26 & 6.1 / 7.4 & 4.7 / 6.3 & 3.95 /3.95 \\
\midrule
\multicolumn{7}{l}{\emph{Steps=512 (NFE=1024); gs=5/gs=6}} \\
0.15 & 11.3 / 12 & 8 / 8.5 & 3.29 / 3.28 & 6.1 / 7 & 4.8 /5.6 & 3.92 / \textbf{3.98} \\
0.19 & 11.6 / 11.5 & 8.2 / 8.3 & 3.31 / 3.26 & 6.1 / 6.6 & 4.5 / 5.2 & 3.94 / 3.95 \\
0.23 & 11.4 / \textbf{10.9} & 8.2 / \textbf{7.7} & 3.27 / 3.25 & 5.6 / 6.4 & 4.2 / 5 & 3.96 / 3.93 \\
0.27 & 11 / 11 & 7.8 / 7.8 & 3.3 / 3.29 & \textbf{5} / 5.6 & \textbf{3.7} / 4 & 3.94 / 3.94 \\
0.31 & 11.3 / 11.3 & 7.8 / 8.1 & 3.31 / 3.27 & 6.2 / 5.6 & 3.7 / 4 & 3.95 / 3.96 \\
0.34 & 15.1 / 11.4 & 12.8 / 8.3 & 3.3 / 3.29 & 5.2 / 5.4 & 4.1 / 4.2 & 3.95 / 3.95 \\
0.37 & 11.4 / 11.4 & \textbf{7.7} / 8 & 3.28 / 3.27 & 6.1 / 5.8 & 4.6 / 4.3 & 3.94 / 3.93 \\
0.4 & 13 / 12.4 & 9.4 / 8.8 & 3.26 / 3.28 & 5.9 / 5.2 & 4.7 / 3.8 & 3.94 / 3.94 \\
\bottomrule
\end{tabular}
\caption{Validation results for text-to-speech transfer, computed on 500 samples from LibriTTS dev-clean. Scores with guidance scales 5 and 6 are reported for each $\gamma$-steps combination.}
\label{tab:audio:val:tts}
\end{table}

\begin{table}[t]
\centering
\footnotesize
\setlength{\tabcolsep}{4pt}
\begin{tabular}{l rrrr}
\toprule
 & \multicolumn{2}{c}{\textbf{CoBit-LibriTTS}} & \multicolumn{2}{c}{\textbf{CoBit-MLS}} \\
\cmidrule(lr){2-3} \cmidrule(lr){4-5}
$\gamma$ & \textbf{WER}$\downarrow$ & \textbf{CER}$\downarrow$  & \textbf{WER}$\downarrow$ & \textbf{CER}$\downarrow$ \\
\midrule
\multicolumn{5}{l}{\emph{Steps=256 (NFE=512); gs=5/gs=6}} \\
0.11 & 20.1 / 20.5 & 13.5 / 13.5 & 8.2 / 8.2 & 4.8 / 4.9 \\
0.13 & 20.6 / 20.9 & 14 / 13.8 & 8.1 / 8.1 & 4.9 / 5 \\
0.15 & 20.2 / 20.6 & 13.5 / 13.8 & 8.3 / 8.1 & 4.8 / 4.8 \\
0.17 & 20.5 / 20.2 & 13.3 / 13.4 & 8.4 / 8.5 & 4.9 / 4.9 \\
0.19 & 20.9 / 20.6 & 13.9 / 13.5 & 8.2 / 8 & 4.8 / 4.7 \\
0.22 & 20 / 20.5 & 13.6 / 13.3 & 8.5 / 8.7 & 5.1 / 5.1 \\
0.25 & 19.8 / 20.3 & 13 / 13.4 & 8.1 / 8.6 & 5.1 / 5.2 \\
0.28 & 20.2 / 20.4 & 13.8 / 13.7 & 8.2 / 8.3 & 5 /5 \\
\midrule
\multicolumn{5}{l}{\emph{Steps=512 (NFE=1024); gs=5/gs=6}} \\
0.15 & 20.3 / 20.1 & 13 / 13.2 & 8.2 / 8.3 & 4.8 / 4.9 \\
0.19 & 20 / 20.2 & 13.2 / 13.1 & 8.5 / 8.2 & 4.9 / 4.9 \\
0.23 & 20.1 / 20 & 12.9 / 13 & 8.3 / 8.4 & 4.8 /5\\
0.27 & 20 / 20 & 13 / 13.2 & \textbf{7.8} / 8.1 & \textbf{4.6} / 4.9 \\
0.31 & 20.2 / 20.2 & 13 / 13.1 & 8.2 / 8.1 & 4.7 / 4.9 \\
0.34 & 19.8 / 20.2 & 12.8 /12.9 & 8.5 / 8.4 & 4.9 / 5\\
0.37 & \textbf{19.4} / 20.1 & \textbf{12.7} / 13 & 8.3 / 8.2 & 4.8 / 4.8 \\
0.4 & 20.2 / 20.2 & 12.9 / 13.1 & 8.3 / 8.5 & 4.9 / 4.9 \\
\bottomrule
\end{tabular}
\caption{Validation results for speech-to-text transfer, computed on 500 samples from LibriTTS dev-clean. Scores with guidance scales 5 and 6 are reported for each $\gamma$-steps combination.}
\label{tab:audio:val:stt}
\end{table}

\begin{table}[t]
\centering
\footnotesize
\setlength{\tabcolsep}{4pt}
\begin{tabular}{l rrrrrrrr}
\toprule
 & \multicolumn{4}{c}{\textbf{CoBit-LibriTTS}} & \multicolumn{4}{c}{\textbf{CoBit-MLS}} \\
\cmidrule(lr){2-5} \cmidrule(lr){6-9}
$\gamma$ & \textbf{WER}$\downarrow$ & \textbf{CER}$\downarrow$ & \textbf{UTMOS}$\uparrow$ & \makecell{\textbf{GenPPL-} \\ \textbf{Speech}$\downarrow$} & \textbf{WER}$\downarrow$ & \textbf{CER}$\downarrow$ & \textbf{UTMOS}$\uparrow$ & \makecell{\textbf{GenPPL-} \\ \textbf{Speech}$\downarrow$} \\
\midrule
\multicolumn{7}{l}{\emph{Steps=256 (NFE=512); gs=5/gs=6}} \\
0.11 & 24.7 / 25.4 & 16.9 / 17.5 & 2.48 / 2.4 & 138.1 / 123.4 & 5.3 / 5.6 & 2.9 / 3.2 & 3.89 / 3.93 & 77.9 / 79.5 \\
0.13 & 23.7 / 23.8 & 16 / 16.2 & 2.48 / 2.42 & 122.7 / 121.5 & 5.6 / 5.2 & 3.4 / 2.9 & 3.89 / 3.91 & 77.1 / 76 \\
0.15 & 23.8 / 23.5 & 16.1 / 16.4 & 2.46 / 2.43 & 123.7 / 116.4 & 5.6 / 5.6 & 3.1 / 3.3 & 3.91 / 3.91 & 76 / 75.4 \\
0.17 & 22.2 / 23.3 & 15 / 16 & 2.46 / 2.41 & 117.9 / 117.6 & 5.5 / 5.4 & 3 / 3.1 & 3.89 / 3.89 & 78.2 / 74 \\
0.19 & 22.3 / 23.5 & 15.5 / 16.1 & 2.47 / 2.4 & 117.4 / 121.1 & 5.4 / 5.5 & 3.1 / 3 & 3.89 / 3.9 & 77.4 / 71.4 \\
0.22 & 22.6 / 22.8 & 15.8 / 15.9 & 2.45 / 2.41 & 112.7 / 110.9 & 6.3 / 5.8 & 3.8 / 3.4 & 3.89 / \textbf{3.94} & 76.3 / 74.5 \\
0.25 & 22.6 / 24.2 & 15.5 / 17 & 2.45 / 2.42 & 112.3 / 108.2 & 5.2 / 5.6 & 3 / 3.1 & 3.9 / 3.89 & 71.5 / 70.3 \\
0.28 & 22.4 / 24.4 & 15.5 / 18.7 & 2.45 / 2.39 & 106.9 / 107.3 & 6 / 5 & 3.7 / 2.8 & 3.92 / 3.91 & 73.6 / 70 \\
\midrule
\multicolumn{7}{l}{\emph{Steps=512 (NFE=1024); gs=5/gs=6}} \\
0.15 & \textbf{19} / 22.1 & \textbf{12.8} / 15.3 & \textbf{2.51} / 2.45 & 102.8 / 100.7 & 5.4 / 5.5 & 3.1 / 3.4 & 3.93 / 3.91 & 70.6 / 67.5 \\
0.19 & 19.9 / 22 & 13.4 / 15.5 & \textbf{2.51} / 2.45 & 100.5 / 101.6 & 5.2 / 5.3 & 2.9 / 3.1 & 3.92 / 3.91 & 67 / 65.9 \\
0.23 & 21 / 22.6 & 16.7 / 15.5 & \textbf{2.51} / 2.44 & 93.5 / 92.7 & 7.1 / 5.5 & 3.9 / 3.4 & 3.92 / 3.9 & 66.8 / 64.5 \\
0.27 & 22.5 / 23.2 & 15.9 / 16.7 & 2.49 / 2.46 & 93.7 / 92.7 & 4.9 / 5.4 & 2.8 / 3 & 3.9 / 3.89 & 66 / 66.6 \\
0.31 & 20.9 / 23.6 & 14.7 / 16.7 & 2.47 / 2.43 & 87.8 / 93.5 & 5.3 / 5.7 & 3 / 3.4 & 3.91 / 3.89 & 63.7 / 61.9 \\
0.34 & 21.2 / 25.1 & 15.2 / 18.5 & 2.47 / 2.42 & 89 / 87.7 & 5.6 / 5 & 3.3 / 2.8 & 3.91 / 3.9 & 63.4 / 58.8 \\
0.37 & 23.5 / 25 & 17 / 18.1 & 2.46 / 2.42 & 84 / 86.7 & 5.3 / 6 & 3.2 / 3.7 & 3.9 / 3.9 & 63.4 / 61.5 \\
0.4 & 21.7 / 24.6 & 15.3 / 17.8 & 2.43 / 2.39 & \textbf{82.8} / 86.2 & \textbf{4.9} / \textbf{4.9} & \textbf{2.7} / 2.8 & 3.9 / 3.91 & 61.3 / \textbf{58.7} \\
\bottomrule
\end{tabular}
\caption{Validation results for speech continuation, computed on 500 samples from LibriTTS dev-clean. Scores with guidance scales 5 and 6 are reported for each $\gamma$-steps combination.}
\label{tab:audio:val:cont}
\end{table}

\subsection{Complete Results} \label{app:audio:results}

\begin{figure}[ht]
    \centering
    \includegraphics[width=\linewidth]{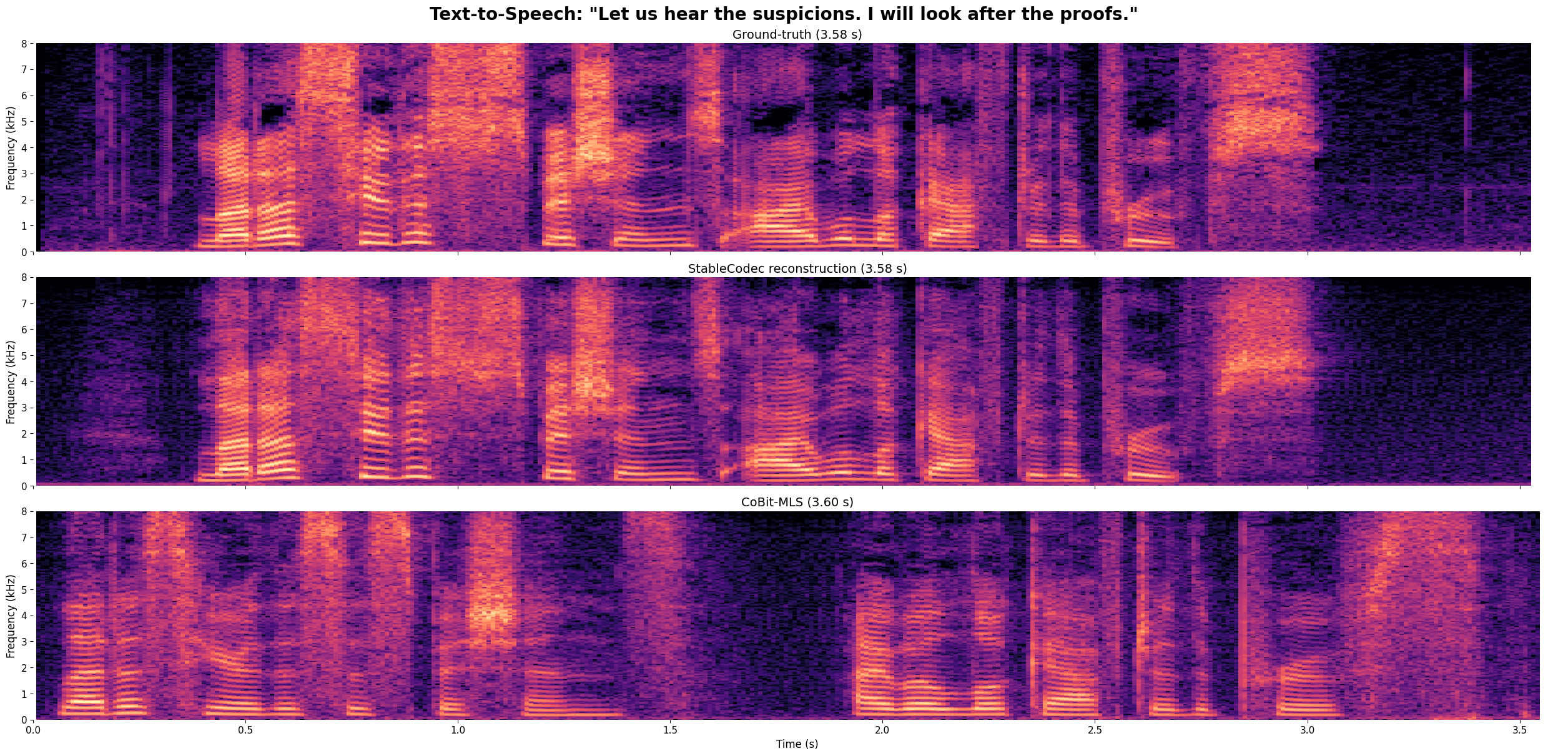}
    \caption{Text-to-speech generation with CoBit-MLS. Given text and speaker tokens (extracted from a held-out example), the model is tasked with generating the speech part of the sequence. The top mel-spectrogram corresponds to the ground truth waveform, the middle one to the StableCodec reconstruction, and the bottom mel-spectrogram to the generated audio by the 633M parameter bitstream diffusion model. The tokenizer introduces visible artifacts, especially at high frequencies. CoBit generation spreads the utterance across a longer time span, without leading and trailing silence, replicating similar frequency patterns. The only exception is the gap between 1.7 and 1.9 seconds, which corresponds to the boundary between sentences.}
    \label{fig:audio:tts-spectrogram}
\end{figure}

\begin{figure} [ht]
    \centering
    \includegraphics[width=\linewidth]{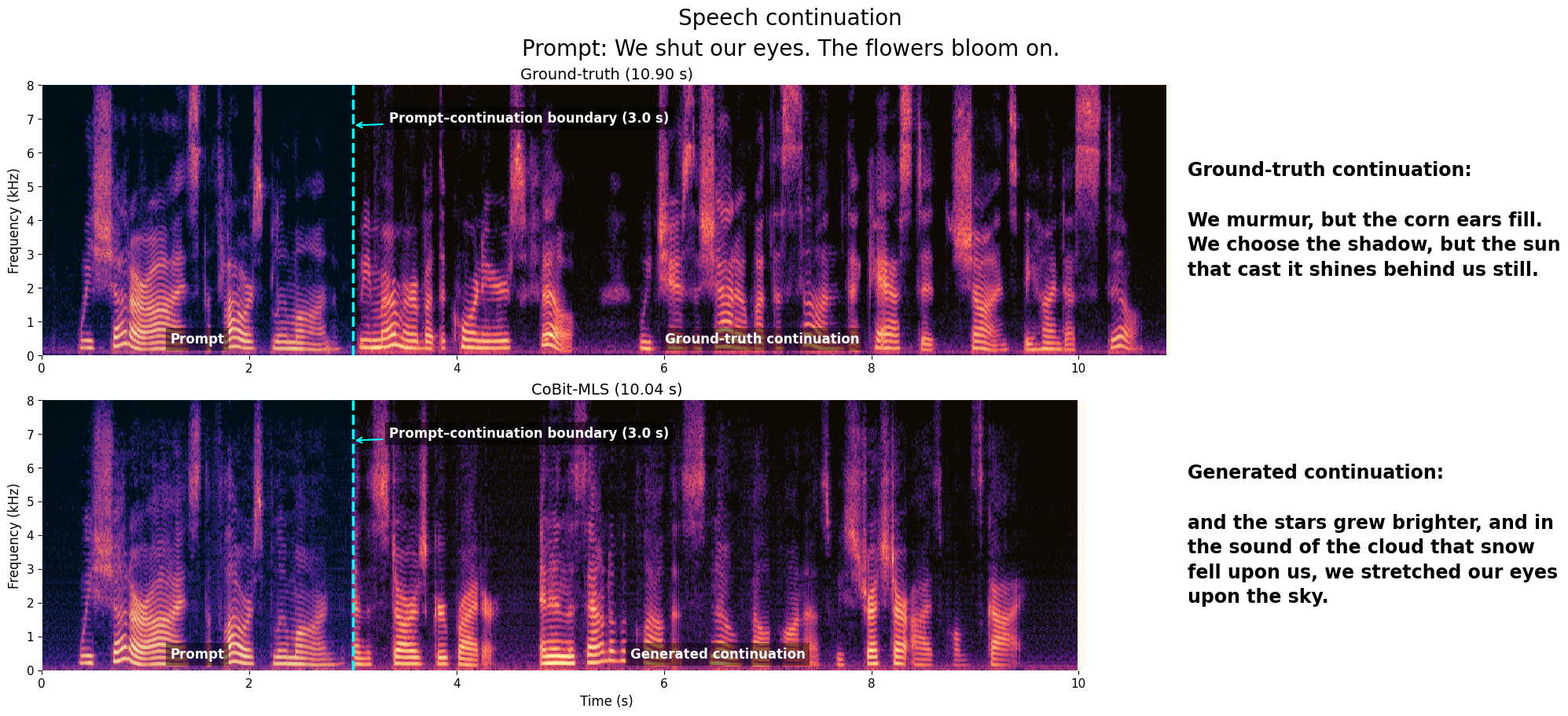}
    \caption{Speech continuation with CoBit-MLS. Given speaker and speech tokens extracted from a 3-second audio prompt, the model is tasked with generating its own continuation. The top mel-spectrogram corresponds to the ground truth continuation, and the bottom mel-spectrogram to the continuation generated by the 633M parameter bitstream diffusion model. The generated spectrogram blends well with the audio prompt, introducing minimal artifacts at the prompt-continuation boundary. Speech continuation is scored not only in terms of acoustic consistency, but also for the semantic relevance and intelligibility of the generated content, shown on the right.}
    \label{fig:audio:continuation-spectrogram}
\end{figure}

Tables \ref{tab:audio:joint}, \ref{tab:audio:cross-modal} and \ref{tab:audio:continuation} contain extensive evaluation metrics and baseline comparisons for co-generation, cross-modal directions and continuation, respectively; Table \ref{tab:audio:salmon} details per-task SALMON accuracy. Mel-spectrogram examples for text-to-speech and speech continuation are visualized in Figures \ref{fig:audio:tts-spectrogram} and \ref{fig:audio:continuation-spectrogram}.

\begin{table}[t]
\centering
\small
\setlength{\tabcolsep}{4.5pt}
\begin{tabular}{lrrrrrr}
\toprule
\textbf{Model} & \textbf{Params} & \textbf{WER}$\downarrow$ & \textbf{CER}$\downarrow$ & \textbf{UTMOS}$\uparrow$ & \textbf{GenPPL-Text}$\downarrow$ & \textbf{GenPPL-Speech}$\downarrow$ \\
\midrule
\multicolumn{7}{l}{\emph{CoBit (ours; from scratch, no text pretraining)}} \\
CoBit-LibriTTS & 134M & 19.9 & 13.3 & 3.02 & 509.6 & 377.2 \\
CoBit-MLS & 633M & 3.5 & \textbf{1.8} & 3.77 & 252.4 & \textbf{133.3} \\
\midrule
\multicolumn{7}{l}{\emph{Autoregressive System (SmolLM+Speaker $\rightarrow$ SparkTTS)}} \\
Cascade & 640M & \textbf{3.2} & 3.1 & \textbf{4.23} & \textbf{122} & 1{,}648 \\
\bottomrule
\end{tabular}
\caption{Joint generation performance, computed by generating $1,024$ samples. Speech perplexity for the cascade system is high, as SmolLM produces multilingual sequences or web/code artifacts that are not transferable to speech. We use the same NFE as in Table \ref{tab:audio:headline}; 256 for CoBit-LibriTTS and 512 for CoBit-MLS.}
\label{tab:audio:joint}
\end{table}

\begin{table}[t]
\centering
\small
\setlength{\tabcolsep}{4.5pt}
\begin{tabular}{lrrrrrrr}
\toprule
& & \multicolumn{4}{c}{\textbf{Text-to-Speech}} & \multicolumn{2}{c}{\textbf{Speech-to-Text}} \\
\cmidrule(lr){3-6} \cmidrule(lr){7-8}
\textbf{Model} & \textbf{Params} & \textbf{WER}$\downarrow$ & \textbf{CER}$\downarrow$ & \textbf{UTMOS}$\uparrow$ & \textbf{SpkSim}$\uparrow$ & \makecell{\textbf{WER}$\downarrow$ \\ \textbf{Clean}} & \makecell{\textbf{WER}$\downarrow$ \\ \textbf{Other}} \\
\midrule
\multicolumn{8}{l}{\emph{CoBit (ours; from scratch, no text pretraining)}} \\
CoBit-LibriTTS & 134M & 11.7 & 8.1 & 3.31 & 0.19 & 23.3 & 38 \\
CoBit-MLS & 633M & 5.3 & 4.1 & 4 & 0.27 & 8.2 & 22 \\
\midrule
\multicolumn{8}{l}{\emph{Autoregressive SLMs/Hybrids}} \\
GPA-v1.5 \citep{audio:gpa} & 505M & \textbf{3.1} & \textbf{3.1} & \textbf{4.37} & \textbf{0.4} & \textbf{3.4} & \textbf{6.3} \\
SpiritLM-Base & 7B & --- & 6.7 & --- & --- & 6 & 11 \\
Moshi \citep{audio:moshi} & 7B & 4.7 & --- & --- & --- & 5.7 &  --- \\
\midrule
\emph{StableCodec (reconstruction)} & --- & --- & --- & 4.3 & 0.52 & --- & --- \\
\bottomrule
\end{tabular}
\caption{Text-to-speech and Speech-to-text performance on LibriSpeech-PC. For text-to-speech, evaluation is conducted on the test-clean partition. For each speaker in the dataset, one sample was held out to extract speaker features. Speech-to-text evaluation is conducted on test-clean and test-other partitions separately. As in Table \ref{tab:audio:headline}, Steps$=$512 for both CoBit-LibriTTS and CoBit-MLS.}
\label{tab:audio:cross-modal}
\end{table}

\begin{table}[t]
\centering
\small
\setlength{\tabcolsep}{4.5pt}
\resizebox{\linewidth}{!}{
\begin{tabular}{lrrrrrrr}
\toprule
\textbf{Model} & \textbf{Params} & \textbf{UTMOS}$\uparrow$ & \textbf{FSD-wlm}$\downarrow$ & \textbf{FSD-e2v}$\downarrow$ & \textbf{SpkSim}$\uparrow$ & \makecell{\textbf{GenPPL-} \\ \textbf{Speech}$\downarrow$} & \textbf{GPT-4o}$\uparrow$ \\
\midrule
\multicolumn{7}{l}{\emph{CoBit (ours; from scratch, no text pretraining)}} \\
CoBit-LibriTTS & 134M & 2.82 & 2{,}373 & 40.5 & 0.21 & 152.1 & 1.11 \\
CoBit-MLS & 633M & 3.89 & 2{,}386 & 13 & 0.34 & 83.6 & 1.44 \\
\midrule
\multicolumn{7}{l}{\emph{Trained on MLS-English (same corpus as CoBit-MLS, excluding LibriTTS)}} \\
Flow-SLM-S \citep{audio:flow-slm} & 395M & 2.47 & 3{,}391 & 35.7 & 0.21 & 242.8 & 1.1 \\
Flow-SLM-L & 1.25B & 3.48 & 3{,}476 & 35.4 & 0.41 & 131.7 & 1.57 \\
\midrule
\multicolumn{7}{l}{\emph{Speech-only SLMs}} \\
Flow-SLM-Ext. & 1.25B & 3.47 & 3{,}445.1 & 33 & 0.42 & 122.4 & 1.67 \\
Llama-Mimi \citep{audio:llama-mimi} & 1.3B & 3.18 & 3{,}621 & 14.1 & 0.35 & 98.5 & 1.58 \\
\midrule
\multicolumn{7}{l}{\emph{Text+Speech SLMs}} \\
TASLM-Embed \citep{audio:taste} & 1.3B & \textbf{4.16} & \textbf{1{,}410} & \textbf{9.9} & \textbf{0.49} & \textbf{38.6} & \textbf{3.42} \\
SpiritLM-Base \citep{audio:spirit-lm} & 7B & 3.2 & 7{,}325 & 26.8 & 0.02 & 101.4 & 2 \\
SpiritLM-Expressive & 7B & 3.17 & 6{,}293.6 & 16.9 & 0.08 & 153.2 & 1.62 \\
\midrule
\emph{StableCodec (reconstruction)} & --- & 4.33 & 1{,}191.6 & 1.26 & 0.36 & --- & --- \\
\bottomrule
\end{tabular}
}
\caption{Continuation-based evaluation for text-audio experiments against published SLMs. All figures are evaluated using publicly available checkpoints. GPT-4o refers to LLM-as-a-Judge score with the prompt shown in Table \ref{tab:audio:llm-judge}. We use Steps$=$512 for both CoBit models. The last row reports the tokenizer performance ceiling.}
\label{tab:audio:continuation}
\end{table}

\begin{table}[t]
\centering
\small
\setlength{\tabcolsep}{4.5pt}
\resizebox{\linewidth}{!}{
\begin{tabular}{lrrrrrrr}
\toprule
\textbf{Model} & \textbf{Sentiment} & \textbf{Speaker} & \textbf{Gender} & \textbf{Bg.- Domain} & \textbf{Bg.- Random} & \textbf{Room} & \textbf{Avg.} \\
\midrule
\multicolumn{7}{l}{\emph{CoBit (ours; from scratch, no text pretraining)}} \\
CoBit-LibriTTS & 86 & 88.5 & 88.5 & 66 & 71 & 71 & 78.5 \\
CoBit-MLS & 90.5 & 93 & 93.5 & 73.5 & 80 & 75.5 & 84.3 \\
\midrule
\multicolumn{7}{l}{\emph{Trained on MLS-English (same corpus as CoBit-MLS, excluding LibriTTS)}} \\
Flow-SLM-S & 85 & 87 & 92.5 & 77 & 86 & 69.5 & 82.83 \\
Flow-SLM-L & 92.5 & 99.5 & 99 & 77 & 84.5 & 77 & 88.25 \\
\midrule
\multicolumn{7}{l}{\emph{Speech-only SLMs}} \\
Flow-SLM-Ext. & 93 & 98 & \textbf{100} & 77 & 86.5 & \textbf{85.5} & \textbf{90} \\
Llama-Mimi & 93 & 90 & 95 & \textbf{81} & \textbf{90} & 78.5 & 87.92 \\
\midrule
\multicolumn{7}{l}{\emph{Text+Speech SLMs}} \\
TASLM-Embed & \textbf{95} & \textbf{100} & \textbf{100} & 61.5 & 54.5 & 57.5 & 78 \\
SpiritLM-Base & 47.5 & 52 & 50.5 & 50.5 & 52.5 & 50 & 50.5 \\
SpiritLM-Expressive & 72 & 54 & 56 & 58.5 & 55.5 & 54.5 & 58.41 \\
\midrule
\emph{StableCodec (reconstruction)} & 99.5 & 99 & 100 & 75.5 & 82 & 95 & 91.83 \\
\bottomrule
\end{tabular}
}
\caption{Speech continuation performance across the consistency tasks of the SALMON benchmark for generation-based evaluation. Each model generates a continuation for each prompt, and accuracy scores are computed for each task by comparing the cosine similarity of the generated continuation with the positive and negative examples. Different embedding models are used for each task, specified in Appendix \ref{app:audio:protocol}. Steps$=$512 is used for both CoBit-LibriTTS and CoBit-MLS, with the operating points selected for speech continuation.}
\label{tab:audio:salmon}
\end{table}

\subsection{Sampler Characterisation} \label{app:audio:sampler}

\textbf{Few-step regimes survive only when generating a single modality.} CoBit-MLS step count sweeps for co-generation, cross-modal directions and continuation are reported in Tables \ref{tab:audio:steps:joint}, \ref{tab:audio:steps:cross-modal} and \ref{tab:audio:steps:continuation}, respectively, with fixed stochasticity ($\gamma=0.175$) and guidance (gs=5 for conditional tasks) settings. For text-to-speech, outputs become well-formed from 16 integration steps and approach the model's best performance with 32 steps, particularly on the acoustic elements (UTMOS and SpkSim). Even less compute is required for speech recognition, with strongly sensible transcriptions being produced with as few as 8 steps. Different behavior is observed for joint generation, where all metrics improve near-monotonically with higher NFEs. In few-step regimes, co-generated samples are inaudible; as the step count increases, naturalness improves and the content becomes semantically plausible for both streams, with decreasing perplexity. The same perplexity pattern is observed for speech continuation as well. The model appears to reach an expressiveness optimum at 32 steps, at the expense of semantic degradation. Beyond that point, CoBit continuations already match the tokenizer's speaker similarity ceiling.

\begin{table}[t]
\centering
\small
\setlength{\tabcolsep}{4.5pt}
\begin{tabular}{lrrrrr}
\toprule
\textbf{Steps} & \textbf{WER}$\downarrow$ & \textbf{CER}$\downarrow$ & \textbf{UTMOS}$\uparrow$ & \textbf{GenPPL-Text}$\downarrow$ & \textbf{GenPPL-Speech}$\downarrow$ \\
\midrule
4 & 367 & 250.5 & 1.28 & 20{,}101.9 & 1{,}132.5 \\
8 & 267.8 & 181.4 & 1.31 & 16{,}791.8 & 708.2 \\
16 & 173.4 & 120.5 & 1.51 & 9{,}002.5 & 303.4 \\
24 & 100 & 70.4 & 1.94 & 5{,}377.7 & 440.5 \\
32 & 68 & 47.4 & 2.3 & 3{,}665.2 & 391.4\\
48 & 41.7 & 28.6 & 2.88 & 1{,}857.4 & 359.5\\
64 & 27.8 & 18.5 & 3.12 & 1{,}267 & 313.3 \\
128 & 9.8 & 6.1 & 3.61 & 397.1 & 207.8 \\
256 & 4.8 & 2.6 & 3.75 & 250.4 & 147.2 \\
512 & \textbf{3.5} & \textbf{1.8} & \textbf{3.77} & \textbf{252.4} & \textbf{133.3} \\
\bottomrule
\end{tabular}
\caption{CoBit-MLS joint generation performance at different step counts, computed by generating $1,024$ samples. All rows inject stochasticity at $\gamma=0.175$. Unconditional generation does not use classifier-free guidance, so each integration step corresponds to a single function evaluation.}
\label{tab:audio:steps:joint}
\end{table}

\begin{table}[t]
\centering
\small
\setlength{\tabcolsep}{4.5pt}
\begin{tabular}{lrrrrrr}
\toprule
& \multicolumn{4}{c}{\textbf{Text-to-Speech}} & \multicolumn{2}{c}{\textbf{Speech-to-Text}} \\
\cmidrule(lr){2-5} \cmidrule(lr){6-7}
\textbf{Steps} & \textbf{WER}$\downarrow$ & \textbf{CER}$\downarrow$ & \textbf{UTMOS}$\uparrow$ & \textbf{SpkSim}$\uparrow$ & \makecell{\textbf{WER}$\downarrow$ \\ \textbf{Clean}} & \makecell{\textbf{WER}$\downarrow$ \\ \textbf{Other}} \\
\midrule
4 & 184.2 & 133.3 & 1.51 & 0.05 & 15.4 & 34.9\\
8 & 32 & 21.1 & 3.03 & 0.19 & 9 & 23.9 \\
16 & 6.6 & 4.5 & 3.85 & 0.25 & 8.6 & 23 \\
24 & 6.7 & 4.8 & 3.96 & 0.26 & 8.4 & 22.8 \\
32 & 5.7 & 4 & 3.99 & 0.26 & 8.3 & 22.7 \\
48 & 5.5 & \textbf{3.9} & 4.02 & \textbf{0.27} & 8.3 & 22.9 \\
64 & 6 & 4.1 & \textbf{4.03} & \textbf{0.27} & 8.3 & 22.5 \\
128 & 6.3 & 4.1 & \textbf{4.03} & \textbf{0.27} & \textbf{8.1} & 22.3 \\
256 & 5.4 & 5.2 & \textbf{4.03} & \textbf{0.27} & 8.2 & 22.4 \\
512 & \textbf{5.3} & 4.1 & 4.01 & 0.26 & 8.2 & \textbf{22.2} \\
\bottomrule
\end{tabular}
\caption{CoBit-MLS cross-modal performance at different step counts, computed by generating $1,024$ samples. All rows inject stochasticity at $\gamma=0.175$. Classifier-free guidance with gs=5 is used throughout the table.}
\label{tab:audio:steps:cross-modal}
\end{table}

\begin{table}[t]
\centering
\small
\setlength{\tabcolsep}{4.5pt}
\begin{tabular}{lrrrrrr}
\toprule
\textbf{Steps} & \textbf{UTMOS}$\uparrow$ & \textbf{FSD-wlm}$\downarrow$ & \textbf{FSD-e2v}$\downarrow$ & \textbf{SpkSim}$\uparrow$ & \textbf{GenPPL-Speech}$\downarrow$ & \textbf{SALMON}$\uparrow$ \\
\midrule
4 & 1.44 & 6{,}001 & 156.2 & 0.05 & 375.1 & 74.58 \\
8 & 1.82 & 3{,}409.6 & 42.1 & 0.14 & 262.6 & 80.08 \\
16 & 3.12 & 2{,}236.3 & 8.9 & 0.3 & 232 & 84.41\\
24 & 3.53 & 2{,}176.6 & 7.1 & 0.33 & 200.9 & 84.41 \\
32 & 3.7 & 2{,}117.6 & \textbf{6.8} & 0.34 & 178.3 & 84.25 \\
48 & 3.83 & 2{,}127.4 & 7.5 & 0.34 & 150.6 & 85.08 \\
64 & 3.88 & 2{,}127.4 & 8.1 & \textbf{0.35} & 130.2 & 84.75 \\
128 & 3.9 & 2{,}108.2 & 9.2 & 0.34 & 111.6 & \textbf{85.58} \\
256 & \textbf{3.91} & \textbf{2{,}108.4} & 10.6 & 0.34 & 94.7 & 84.41 \\
512 & \textbf{3.91} & 2{,}179.3 & 11.8 & 0.34 & \textbf{85.8} & 83.92 \\
\bottomrule
\end{tabular}
\caption{CoBit-MLS speech continuation performance at different step counts, computed by generating $1,024$ samples. All rows inject stochasticity at $\gamma=0.175$. Classifier-free guidance with gs=5 is used throughout the table.}
\label{tab:audio:steps:continuation}
\end{table}

\textbf{Stochasticity is crucial for co-generation.} In Table \ref{tab:audio:deterministic}, we examine the use of a deterministic sampler for CoBit-MLS. For conditional tasks, deterioration compared to a stochastic sampler is minimal, and only apparent in the generative perplexity of speech continuation. In fact, text-to-speech CER, FSD-e2v and SALMON accuracy improve. In joint generation, however, stochasticity is essential: alignment loosens with error rates more than doubling, and the content becomes far less intelligible, especially in the text stream.

\begin{table}[t]
\centering
\small
\setlength{\tabcolsep}{4.5pt}
\resizebox{\linewidth}{!}{
\begin{tabular}{lrrrrrrr}
\toprule
\textbf{Direction} & \textbf{WER}$\downarrow$ & \textbf{CER}$\downarrow$ & \textbf{UTMOS}$\uparrow$ & \textbf{SpkSim}$\uparrow$ & \textbf{FSD}$\downarrow$ &  \textbf{GenPPL}$\downarrow$ & \textbf{SALMON}$\uparrow$ \\
\midrule
\multicolumn{8}{l}{\emph{Deterministic Sampler}} \\
text $\rightarrow$ speech & 5.9 & 4.1 & 3.97 & 0.26 & --- & --- & --- \\
speech $\rightarrow$ text & 8.6 / 23.4 & 4.7 / 14 & --- & --- & --- & --- & --- \\
speech continuation & --- & --- & 3.85 & 0.33 & 2{,}130.4 / 8.2 & 136.9 & 85.41 \\
joint & 11.3 & 7.3 & 3.54 & --- & --- & 413.9 / 204.2 & --- \\
\midrule
\multicolumn{8}{l}{\emph{Stochastic sampler, $\gamma$=0.175, gs=5}} \\
text $\rightarrow$ speech & 5.3 & 5.2 & 4.03 & 0.27 & --- & --- & --- \\
speech $\rightarrow$ text & 8.2 / 22.4 & 4.3 / 13 & --- & --- & --- & --- & --- \\
speech continuation & --- & --- & 3.91 & 0.34 & 2{,}108.4 / 10.6 & 94.7 & 84.41 \\
joint & 4.8 & 2.6 & 3.75 & --- & --- & 250.4 / 147.2 & --- \\

\bottomrule
\end{tabular}
}
\caption{CoBit-MLS scores with a deterministic sampler across all four tasks. Integration steps are fixed to 256 and the conditional task use classifier-free guidance with strength 5.}
\label{tab:audio:deterministic}
\end{table}
 
\textbf{FSD deteriorates with guidance.} Tables \ref{tab:audio:guidance:cross-modal} and \ref{tab:audio:guidance:continuation} report guidance sweeps for CoBit-MLS, keeping 256 integration steps and stochasticity $\gamma=0.175$ throughout. While guidance improves modality alignment and generative perplexity, reducing the guidance strength or even shifting to the plain conditional setting improves both FSD metrics, with FSD-e2v surpassing TASLM-Embed for gs=1 and gs=2. Table \ref{tab:audio:guidance:fsd} breaks down FSD calculation. For FSD-wlm, stronger guidance leads to a mean shift; this is natural, as guidance steers generation towards MLS-English, whose pre-processing and frame-level noise may differ substantially from LibriSpeech-PC. For FSD-e2v, only the variance is affected; both datasets consist of reading-style speech and hence contain similar expressiveness.

\begin{table}[t]
\centering
\small
\setlength{\tabcolsep}{4.5pt}
\begin{tabular}{lrrrrrr}
\toprule
& \multicolumn{4}{c}{\textbf{Text-to-Speech}} & \multicolumn{2}{c}{\textbf{Speech-to-Text}} \\
\cmidrule(lr){2-5} \cmidrule(lr){6-7}
\textbf{gs} & \textbf{WER}$\downarrow$ & \textbf{CER}$\downarrow$ & \textbf{UTMOS}$\uparrow$ & \textbf{SpkSim}$\uparrow$ & \makecell{\textbf{WER}$\downarrow$ \\ \textbf{Clean}} & \makecell{\textbf{WER}$\downarrow$ \\ \textbf{Other}} \\
\midrule
1 & 8.6 & 6 & 3.77 & 0.25 & 9.3 & 23.8\\ 
2 & 5.8 & 4.9 & 3.99 & 0.26 & 8.2 & \textbf{22.1} \\
3 & 6.5 & 5.2 & 4.01 & 0.26 & \textbf{8.1} & 22.2 \\
4 & \textbf{5.4} & \textbf{4.5} & \textbf{4.03} & \textbf{0.27} & \textbf{8.1} & 22.2\\
5 & \textbf{5.4} & 5.2 & \textbf{4.03} & \textbf{0.27} & 8.2 & 22.4 \\
\bottomrule
\end{tabular}
\caption{CoBit-MLS cross-modal performance at different guidance scales. All rows use 256 integration steps (256 function evaluations at gs=1; 512 for gs $>$ 1) and stochasticity at $\gamma=0.175$.}
\label{tab:audio:guidance:cross-modal}
\end{table}

\begin{table}[t]
\centering
\small
\setlength{\tabcolsep}{4.5pt}
\begin{tabular}{lrrrrrr}
\toprule
\textbf{gs} & \textbf{UTMOS}$\uparrow$ & \textbf{FSD-wlm}$\downarrow$ & \textbf{FSD-e2v}$\downarrow$ & \textbf{SpkSim}$\uparrow$ & \textbf{GenPPL-Speech}$\downarrow$ & \textbf{SALMON}$\uparrow$ \\
\midrule
1 & 3.88 & \textbf{1{,}953.7} & \textbf{5.1} & 0.33 & 172.1 & 83.41 \\
2 & \textbf{3.94} & 2{,}054.1 & 7.8 & 0.34 & 116 & \textbf{85} \\ 
3 & 3.92 & 2{,}073.1 & 9.1 & 0.34 & 116 & \textbf{85} \\
4 & 3.92 & 2{,}140.3 & 10 & 0.34 & 96.1 & 84.83 \\ 
5 & 3.91 & 2{,}108.4 & 10.6 & 0.34 & \textbf{94.7} & 84.41 \\
\bottomrule
\end{tabular}
\caption{CoBit-MLS speech continuation performance at different guidance scales. All rows use 256 integration steps (256 function evaluations at gs=1; 512 for gs $>$ 1) and stochasticity at $\gamma=0.175$.}
\label{tab:audio:guidance:continuation}
\end{table}

\begin{table}[t]
\centering
\small
\setlength{\tabcolsep}{4.5pt}
\resizebox{\linewidth}{!}{
\begin{tabular}{lrrrrrr}
\toprule
& \multicolumn{3}{c}{\textbf{FSD-wlm}} & \multicolumn{3}{c}{\textbf{FSD-e2v}} \\
\cmidrule(lr){2-4} \cmidrule(lr){5-7}
\textbf{gs} & \textbf{Mean Term} & \textbf{Covariance Term} & \textbf{Total} & \textbf{Mean Term} & \textbf{Covariance Term} & \textbf{Total} \\
\midrule
1 & 684.5 & 1{,}269.1 & 1{,}953.7 & 0.19 & 4.86 & 5.05 \\ 
2 & 748.7 & 1{,}305.4 & 2{,}054.1 & 0.3 & 7.51 & 7.81 \\
3 & 756.3 & 1{,}316.8 & 2{,}073.1 & 0.33 & 8.78 & 9.11 \\
4 & 792.7 & 1{,}347.6 & 2{,}140.3 & 0.36 & 9.65 & 10.01 \\
5 & 777.2 & 1{,}331.3 & 2{,}108.5 & 0.37 & 10.24 & 10.61 \\
\midrule
\emph{StableCodec (reconstruction)} & 238.5 & 953.1 & 1{,}191.6 & 0.07 & 1.19 & 1.26 \\
\bottomrule
\end{tabular}
}
\caption{Breakdown of FSD-wlm and FSD-e2v metrics across the mean term $\|\mu-\mu'\|^2$ and the covariance term $\text{trace}\{\Sigma+\Sigma'-2(\Sigma\Sigma')^\frac{1}{2}\}$ for guidance sweeps. All rows use 256 integration steps (256 function evaluations at gs=1; 512 for gs $>$ 1) and stochasticity at $\gamma=0.175$.}
\label{tab:audio:guidance:fsd}
\end{table}

\subsection{Effect of Model and Corpus Scale} \label{app:audio:scale}

Increasing the model scale and adding MLS-English to the training set improves performance across all tasks and metrics. The only exception is FSD-wlm, where CoBit-LibriTTS benefits from its training set being a subset of the LibriSpeech-PC training partition, and is threrefore more similar to the reference population. MLS-English is much larger (45k hours versus 580), preserves reading-style audio and is part of StableCodec's training corpus as well, reducing the risk of distributional shifts during tokenization. CoBit-MLS is better at aligning text and speech sequences during co-generation, yet CoBit-LibriTTS achieves its best error rates when given the text as input. The performance gap is small for SALMON accuracy, as the smaller model can still capture acoustic properties of the speech prompt. Despite its linguistic fluency and frame-level realism, the main limitation of CoBit-LibriTTS is producing natural-sounding audio, observed across all speech-emitting tasks in terms of UTMOS. In generative perplexity, UTMOS, FSD-wlm and SpkSim for speech continuation, CoBit-LibriTTS matches or even outperforms the 395M parameter Flow-SLM version. For LLM-as-a-Judge, however, CoBit-LibriTTS produces almost exclusively irrelevant continuations.  In SALMON, similar to CoBit-MLS, it scores right between multimodal autoregressive SLMs and speech-only models, except its average accuracy is lower than that of Flow-SLM-S.

\subsection{Speech tokenization and Speaker Features} \label{app:audio:speech-tokenization}

\textbf{StableCodec performance ceiling.} The last rows of Tables \ref{tab:audio:cross-modal}, \ref{tab:audio:continuation} and \ref{tab:audio:salmon} report the evaluation of test set reconstruction with the StableCodec configuration used throughout the text-audio experiments, for all metrics where tokenization determines the best obtainable performance. StableCodec alters the speaker identity and frame-level distribution of the test set substantially, restricting the SpkSim and FSD-wlm scores the model can reach. For speech continuation, CoBit-MLS matches the SpkSim ceiling, while a large portion of its reported FSD-wlm is related to the tokenizer's fidelity. StableCodec's limitated ability to reproduce fine-grained details to the same extent as other tokenizers, such as Mimi, is further observed in the SALMON background-related tasks, where StableCodec reconstruction receives even worse accuracy than all Flow-SLM and Llama-Mimi continuations.

\textbf{Training-inference mismatch.} For all three conditional tasks, BiCodec-global speaker tokens are part of the condition. For ASR, speaker tokens are extracted from the full speech signal, which is always provided as input. Inference-time speaker tokens are extracted from the audio prompt for continuation and from a held-out recording for TTS to rule out information leakage. This is standard practice for zero-shot TTS. During training, however, speaker features from the full ground-truth audio samples are used for all tasks. This forces CoBit to learn a tighter relationship between speaker features and speech tokens, simultaneously accounting for the variability in BiCodec-global, which does not guarantee that all recordings of the same speaker produce identical outputs. A training-inference mismatch is therefore introduced to facilitate unified training. Table \ref{tab:audio:spksim} documents the change in performance if the same type of ground-truth speaker tokens are used during inference (note that this is conducted solely to quantify the effect of ground-truth speaker tokens; it includes information leakage and cannot be considered a valid evaluation for CoBit). Speaker similarity increases from 0.27 to 0.32, still well below GPA-v1.5. The held-out samples themselves are much more similar to the ground truth test examples, with SpkSim well beyond StableCodec's ceiling.

\begin{table}[t]
\centering
\small
\setlength{\tabcolsep}{4.5pt}
\begin{tabular}{lrrrr}
\toprule
 & \textbf{WER}$\downarrow$ & \textbf{CER}$\downarrow$ & \textbf{UTMOS}$\uparrow$ & \textbf{SpkSim}$\uparrow$ \\
\midrule
\multicolumn{5}{l}{\emph{CoBit-MLS, Steps=512, $\gamma=0.27$, gs=5}} \\
held-out features & 5.3 & 4.1 & 4 & 0.27 \\
ground-truth features & 5.2 & 3.6 & 4.03 & 0.32 \\
\midrule
\emph{Held-out vs. ground truth samples} & --- & --- & --- & 0.65 \\
\bottomrule
\end{tabular}
\caption{Effect of speaker tokens on text-to-speech performance. Both CoBit-MLS rows are obtained from text-to-speech generation.}
\label{tab:audio:spksim}
\end{table}

\begin{figure}[ht]
    \centering
    \includegraphics[width=\linewidth]{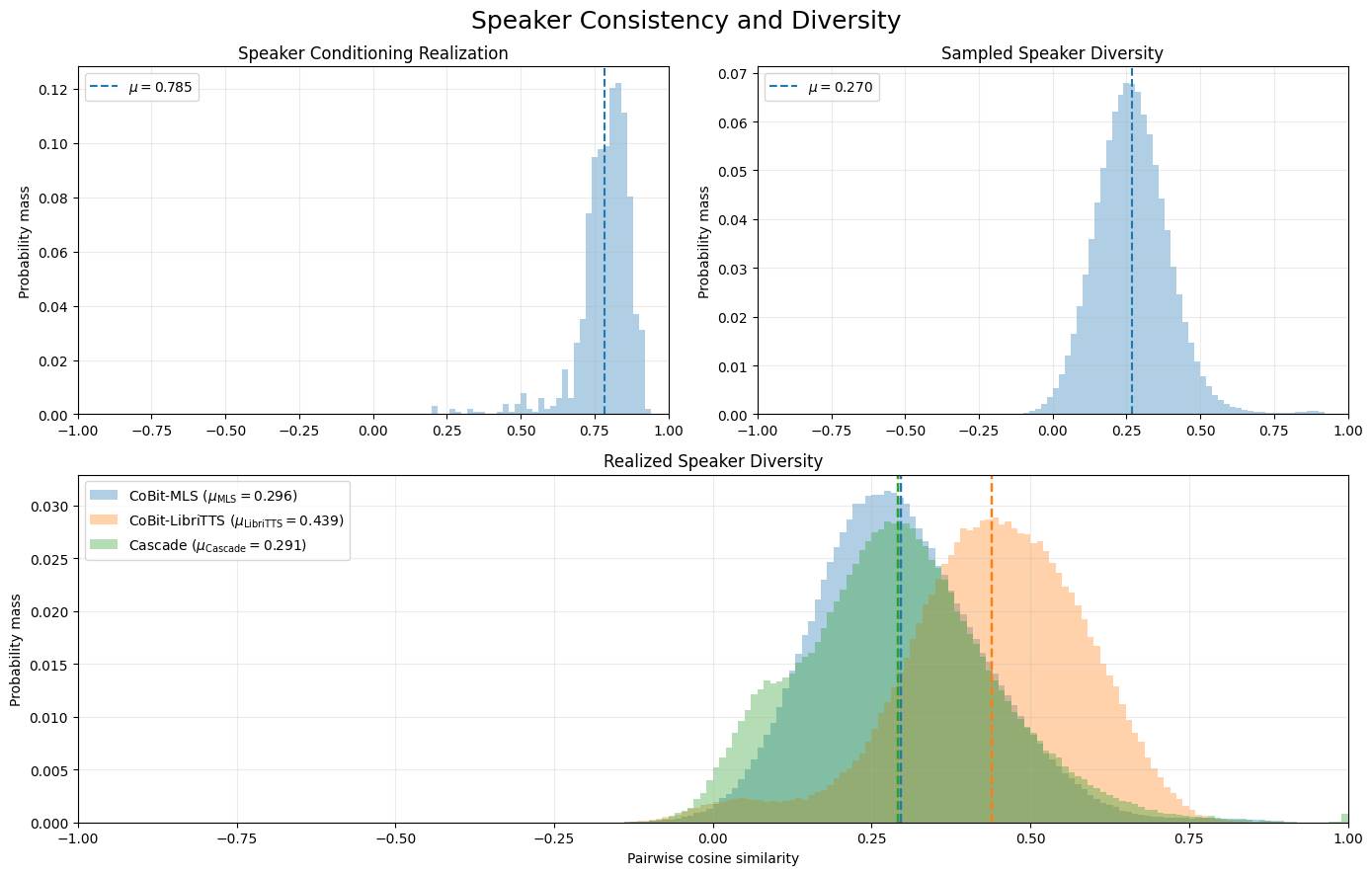}
    \caption{Speaker consistency and diversity distribution for co-generation, in terms of latent space cosine similarity. Top left: Per-sample embedding similarity between BiCodec-global FSQ components sampled by CoBit-MLS (Steps=512, $\gamma=0.175$) during co-generation and BiCodec-global FSQ components extracted from the generated waveforms (decoded StableCodec tokens). Top right: Embedding similarity of BiCodec-global FSQ components across the 1{,}024 BiCodec-global sequences sampled by CoBit-MLS. Both plots are averaged over sequence positions. Bottom: realized speaker diversity among the generated populations of CoBit-MLS, CoBit-LibriTTS (Steps=256, $\gamma=0.175$) and the SmolLM-Speaker-SparkTTS autoregressive cascade (Appendix \ref{app:audio:baselines}), measured in terms of embeddings from the speaker verification CAM++ checkpoint.}
    \label{fig:audio:speaker}
\end{figure}

\textbf{Co-generation: Speaker Realization.} During joint generation, CoBit samples speaker tokens, yet it does not use them for waveform decoding. To measure how speaker conditioning is realized to audio, we compare the sampled speaker tokens to the ones BiCodec-global extracts from the decoded waveform for CoBit-MLS; results are shown in Table \ref{tab:audio:realized-conditioning}. Sampled and realized token sequences agree on 7.62\% of the total tokens, which corresponds to 2-3 tokens per example. Since BiCodec-global is FSQ-based, disagreement in a single latent dimension produces an entirely different token, and thus the token agreement rate is highly sensitive to StableCodec artifacts, that, as already discussed for SpkSim, are recognizable for ECAPA-TDNN, which BiCodec-global is also based on. For this reason, we also examine the relationship between latent BiCodec-global representations that correspond to the FSQ codes, in terms of RMSE and cosine similarity (because FSQ bounds each latent dimension, RMSE can be normalized with respect to the highest attainable value). Sampled and realized continuous latents are closely aligned, showing that CoBit-MLS learns to realize its speaker conditioning, despite not reproducing exact BiCodec-global sequences. The top left panel of Figure \ref{fig:audio:speaker} also verifies that this behavior is observed for the majority of the samples and is not driven by outliers.  

\begin{table}[t]
\centering
\small
\setlength{\tabcolsep}{4.5pt}
\begin{tabular}{lrrrr}
\toprule
& \textbf{Token Agreement(\%)} & \textbf{Position RMSE(\%)} & \textbf{Sequence RMSE(\%)} & \textbf{Cosine Similarity}\\
\midrule
Paired & 7.62 & 5.08 & 5.69 & 0.785 \\
Shuffled & 0.26 ($\pm$0.03) & 10.58 ($\pm$0.03) & 10.99 ($\pm$0.03) & 0.222 ($\pm$0.005) \\
\bottomrule
\end{tabular}
\caption{Speaker conditioning realization for CoBit-MLS co-generation. Comparisons are drawn between BiCodec-global representations CoBit-MLS samples directly (speaker positions in its generated sequence) and BiCodec-global representations extracted from the waveforms CoBit-MLS generates. RMSE and cosine similarity metrics are computed on BiCodec-global's latent space by transforming each FSQ code into its underlying latent vector. The first row computes paired sample correspondence (generated and realized representations for the same sequence); the second row gives a shuffled baseline for shuffled examples, averaged over 100 draws. Cosine similarity is averaged across 32 sequence positions and 1024 samples.}
\label{tab:audio:realized-conditioning}
\end{table}

\textbf{Co-generation: speaker diversity.} For co-generation, we are also interested in whether CoBit can actually learn to draw random voices, instead of repeating a small set of identities. To test this, we calculate the similarity among generated populations under an external speaker verification system that is unrelated to ECAPA-TDNN and BiCodec, namely CAM++ \citep{audio:campp}. We compare CoBit-LibriTTS and CoBit-MLS with the cascaded joint generation pipeline (Appendix \ref{app:audio:baselines}), which learns to autoregressively sample speaker tokens using the same training set and representation as CoBit-MLS and realizes speaker conditioning via the BiCodec decoder. Results are reported in Table \ref{tab:audio:diversity} and the distributions of within-distribution pairwise cosine similarity are visualized in the bottom panel of Figure \ref{fig:audio:speaker}. CoBit-MLS generates as diverse speakers as the baseline on average, with significant overlap in their similarity distributions; CoBit-LibriTTS, having been trained on a much narrower speaker collection, is heavily limited in the range of voices it can reproduce.

\begin{table}[t]
\centering
\footnotesize
\setlength{\tabcolsep}{4pt}
\begin{tabular}{lrrr}
\toprule
& \multicolumn{2}{c}{\textbf{BiCodec-global}} & \textbf{CAM++} \\
\cmidrule(lr){2-3} \cmidrule(lr){4-4}
\textbf{Model} & \textbf{Entropy(bits)} & \textbf{Cosine Similarity (sampled)} & \textbf{Cosine Similarity (realized)} \\
\midrule
\multicolumn{4}{l}{\emph{CoBit (ours; from scratch, no text pretraining)}} \\
CoBit-LibriTTS & --- & --- & 0.439 \\
CoBit-MLS & 8.64 & 0.27 & 0.295 \\
\midrule
\multicolumn{4}{l}{\emph{Autoregressive System (SmolLM+Speaker $\rightarrow$ SparkTTS)}} \\
Cascade & --- & --- & 0.29 \\
\midrule
\multicolumn{4}{l}{\emph{LibriTTS+MLS-English}} \\
Matched Population & 8.75 ($\pm$0.01) & 0.3 ($\pm$0.002) & --- \\
Full & 9.97 & --- & ---\\
\bottomrule
\end{tabular}
\caption{Co-generation speaker diversity, measured in terms of speaker-token entropy and cosine similarity of the sampled BiCodec-global latents and the extracted CAM++ embeddings. Matched population statistics for the training set are computed using 100 random draws.}
\label{tab:audio:diversity}
\end{table}

\begin{figure}[ht]
    \centering
    \includegraphics[width=\linewidth]{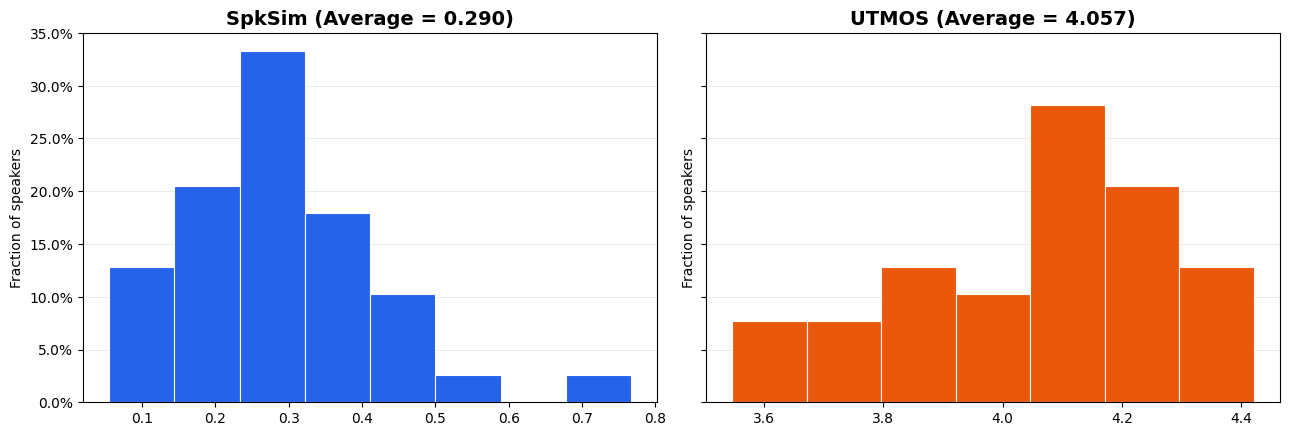}
    \caption{(Left) SpkSim and (Right) UTMOS score distributions for text-to-speech for the 39 speakers in LibriSpeech-PC test-clean. For 10 text prompts, we generate audio with CoBit-MLS using held-out speaker tokens from all speakers in the test set. We use Steps=512, $\gamma=0.27$ and gs=5, as selected for standard text-to-speech evaluation.}
    \label{fig:audio:tts-vc}
\end{figure}

\textbf{Text-to-speech: speaker realization.} We also evaluate the model's ability to realize different voices in text-to-speech transfer, by generating audio for the same 10 text prompts with all 39 speakers of LibriSpeech-PC test-clean, using CoBit-MLS. Figure \ref{fig:audio:tts-vc} shows how SpkSim and UTMOS scores are distributed across the different speakers. For some speakers, the BiCodec-global representation translates better than others, with speaker similarities ranging from 0.1 to 0.8; UTMOS values further verify samples for all speakers maintain natural sound.

\subsection{Speech Recognition Error Breakdown} \label{app:audio:asr}

\begin{figure}[ht]
    \centering
    \includegraphics[width=\linewidth]{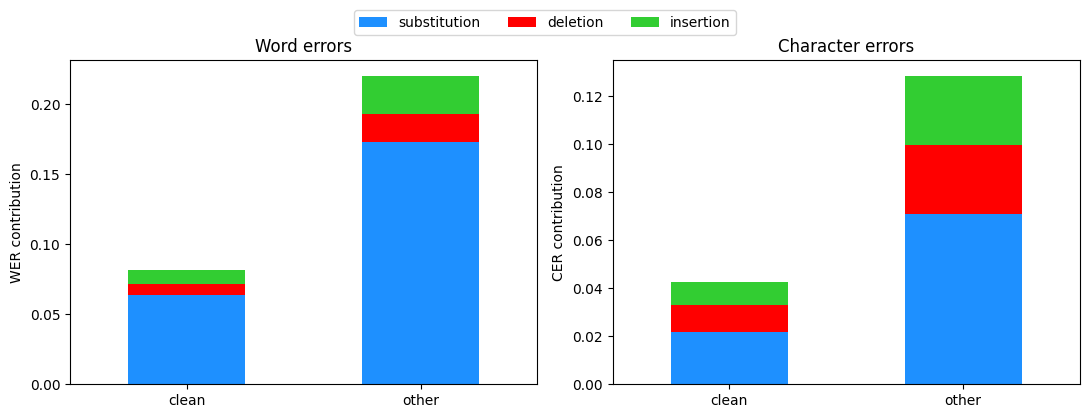}
    \caption{Breakdown of speech-to-text WER and CER for CoBit-MLS, using the selected setting in Table \ref{tab:audio:operating-points}.}
    \label{fig:audio:asr}
\end{figure}

Speech-to-text is the task where cold initialization is the most apparent, as it contradicts the design of ASR-specific systems, which typically recognize speech at character, token, or phoneme level, and then employ a text language model to decode the output into well-structured sentences. For SLMs, the textually pretrained transformer trunk implicitly plays a similar role, leading to more coherent transcriptions, as previously demonstrated by \cite{audio:voxtlm}. Figure \ref{fig:audio:asr} separates CoBit-MLS speech-to-text errors into substitutions, insertions and deletions. The vast majority of word-level errors are substitutions: the model very rarely hallucinates additional words or skips existing ones; it just lacks the language modeling capacity to leverage character- and word-level context to achieve more accurate recognition.

\subsection{LLM-as-a-Judge Score Distributions} \label{app:audio:llm}

\begin{figure}[ht]
    \centering
    \includegraphics[width=\linewidth]{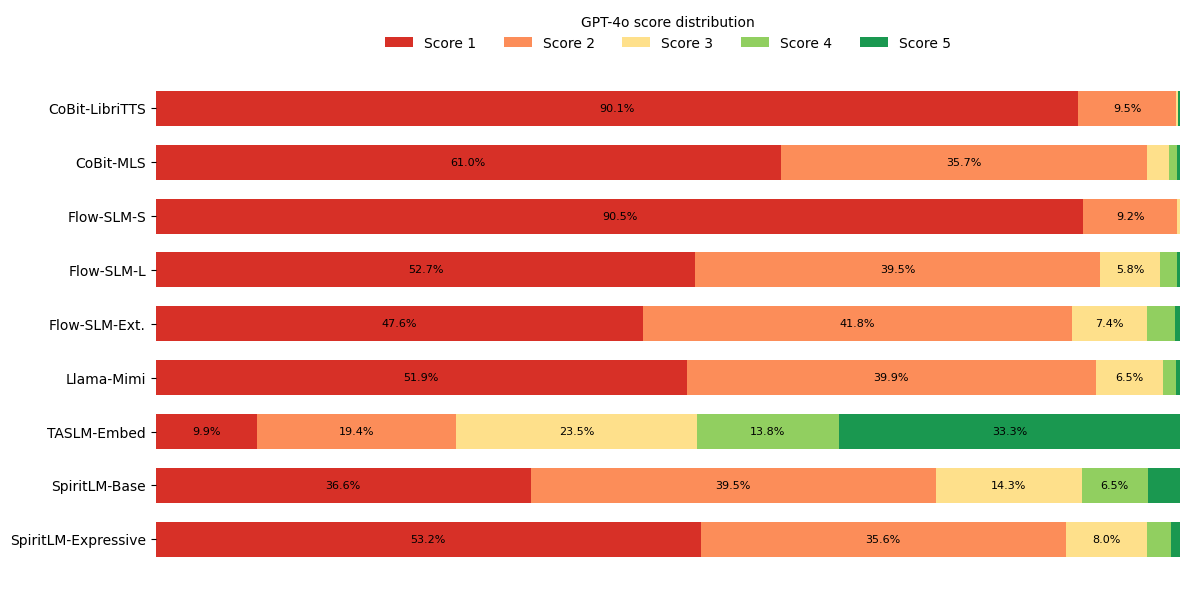}
    \caption{LLM-as-a-Judge score distribution for CoBit and autoregressive SLMs. CoBit-LibriTTS almost always receives the lowest possible score. A large portion of CoBit-MLS samples perform moderately better, but neither of the model can consistently achieve scores of 3 or higher.}
    \label{fig:audio:llm-judge}
\end{figure}

Per-model distributions of LLM-as-a-Judge scores are shown in Figure \ref{fig:audio:llm-judge}. CoBit-MLS produces relevant continuations 39\% of the time, yet even then relevance to the prompt's theme is only marginal. CoBit-LibriTTS produces almost exclusively irrelevant continuations. Nonetheless, both models score on average higher than Flow-SLM-S, which receives the worst score of any studied SLM. CoBit is trained from scratch and only sees storybook passages, whose content is very difficult to generalize, even to other stories. Text pretraining effectively exposes SLMs to much richer semantics, as text-only corpora are typically more diverse. Even warm initialization, however, appears to be insufficient for semantic extrapolation; thematically accurate speech is only achieved by TASLM-Embed, which uses text-aligned speech representations that close the semantic gap between text and speech modeling.

\subsection{Baseline Systems} \label{app:audio:baselines}

The SLM baselines considered in this work are autoregressive; diffusion models in speech generation are mostly limited to TTS systems or ASR-TTS hybrids. 

\textbf{Multimodal SLMs and Hybrids.} The most comparable text-audio baselines for CoBit are models that learn to generate both text and speech. TASLM \citep{audio:taste} generates the two modalities jointly, leveraging continuous text-aligned speech representations from the TASTE encoder and implementing separate prediction heads. The semantic equivalence of text units and speech embeddings bridges the gap between audio and text generation, producing linguistically fluent and conceptually consistent audio. Other approaches focus on systems that can be extended to cross-modal tasks. SpiritLM \citep{audio:spirit-lm} is trained on interleaved text-audio sequences, and has an expressive variant with additional control over pitch and style. SpiritLM models can perform TTS and ASR via task-specific fine-tuning and few-shot prompting. Moshi \citep{audio:moshi} is a spoken dialogue system that produces text alongside speech as an inner monologue that steers generation towards intelligible sentences. Its cross-modal capabilities arise from a time delay on the generated stream and inference-time teacher-forcing on the conditioning stream. GPA \citep{audio:gpa} is a multi-task autoregressive hybrid that treats TTS and ASR as conditional language modelling problems. It is the closest method to CoBit in terms of scale, multi-task joint training and initialization from scratch, but it is limited to cross-conditional tasks and cannot perform speech language modelling. However, all of these models require task-specific design and modifications for TTS and ASR, in contrast to CoBit.

\textbf{Speech-only SLMs.} In the context of speech generation, additional comparisons can be drawn to textless SLMs, which mainly rely on the powerful Mimi tokenizer to balance semantic information with fine-grained acoustic detail. Llama-Mimi \citep{audio:llama-mimi} autoregressively predicts each Mimi layer, leveraging RVQ's conditional nature: the generation of each code is conditioned on both the previous tokens and the earlier codebooks of the same token. Flow-SLM \citep{audio:flow-slm} exploits Mimi's semantic-acoustic combination for cascaded prediction, breaking the strictly autoregressive paradigm: a causal transformer predicts the semantically distilled first RVQ layer; then, unquantized Mimi embeddings are inferred with flow matching, conditional on look-ahead semantic tokens.

\textbf{Cascaded Co-generation.} None of the multimodal systems are evaluated on unconditional co-generation. TASLM-Embed and Moshi are the only baselines that are, in theory, capable of true parallel generation, but they are not trained for such a task. Matching the autoregressive nature of all other baselines, we instead compare CoBit to a cascaded pipeline: text is generated unconditionally with a text LM and speaker identity is sampled from a small transformer; a TTS system then uses the text and speaker features to synthesize the speech sequence. This cascade system effectively targets the same distribution as CoBit for joint generation, factorized as $p(\text{text, speaker, speech}) = p(\text{text})p(\text{speaker})p(\text{speech}|\text{text, speaker})$. To match the $630$M parameter scale of CoBit, SmolLM2-135M \citep{audio:smollm} is used for text generation and Spark-TTS \citep{audio:spark-tts} is used for speech synthesis. Speaker features are sampled by a transformer with 1M parameters in the form of BiCodec-global tokens and passed to the Spark-TTS decoder. We train the speaker transformer on LibriTTS and MLS-English. All three components of the cascade are autoregressive. A known drawback is that unconditional generation with a generic text LM can lead to web or code artifacts that cannot be synthesized naturally as speech, causing very high GenPPL-Speech, as observed in Table \ref{tab:audio:joint}.

\subsection{Measurement Comparability} \label{app:audio:comparability}

For speech continuation and co-generation, we evaluate all baselines under the same protocol as CoBit using publicly available checkpoints. For cross-modal transfer, we only evaluate GPA-v1.5. While public SpiritLM and Moshi models do exist, their TTS and ASR performance relies on task-specific fine-tuning and engineering modifications (\ref{app:audio:baselines}) that cannot be reproduced exactly from public checkpoints; we therefore report results from their respective papers. All generation-based SALMON scores are taken directly from \cite{audio:gen-salmon}. However, the exact protocol documented in the original work is replicated, using the same checkpoints for the judge models.
 
\FloatBarrier
\section{Protein sequence and structure experiments}
\label{app:protein}
\begingroup
\raggedbottom

This appendix documents the checkpoint, sampling configuration and evaluations reported in \S\ref{sec:exp:protein}. All joint-generation settings use 100 pairs, with 20 at each length in $\{100,200,300,400,500\}$.

\subsection{Data and splits}
\label{app:protein:protocol}

We begin by merging DPLM-2 train and validation groups and obtaining its structural codes before constructing our own partitions from the PDB/AFDB-SwissProt sequence--structure pairs \citep{protein:dplm2, protein:pdb, protein:afdb}. For consistency and correctness, we reject empty sequences, unparseable codes and sequence/code length mismatches. To keep the a comparable protocol, long chains retain their first 512 residues. Segments with over 10\% noncanonical residues are removed. Remaining noncanonical residues and out-of-range codes are masked. Despite strong validation checks, only seven records have been removed and $21{,}574$ were truncated. No additional resolution or pLDDT filter is applied. 

Whole clusters with sequences were assigned deterministically to training or validation. We reserve 128 CAMEO/CASP-labelled sequences and hash the remaining cluster IDs with SHA-256, using a 2\% validation threshold. This gives us $216{,}237$ training and $4{,}110$ validation records with disjoint cluster IDs. This was done before the training loader's 40--512 residues restrictions were applied. PDB:AFDB sampling ratio is 2:1. 

\textbf{Sequence replay.} The March-2020 EvoDiff UniRef50 training partition supplied $400{,}000$ sequences \citep{protein:uniref}. A NumPy Generator (seed 0) permutes source indices and chooses fixed contiguous 512-residue crops for the long chains. Replay supplies $20\%$ of training batches with structure channels masked.

\textbf{Evaluation sets.} We use established validation sets in our runs. For CATH~4.3 topology validation \citep{protein:cath} we retain each chain's longest contiguous segment with finite N/C$\alpha$/C/O coordinates, canonical residues. NumPy (seed 20260827) selects 128 of the $1{,524}$ eligible chains without replacement. Indices are sorted. These targets also determine sampler selection, so the scores are validation results. Forward folding uses all 183 DPLM-2.1 CAMEO~2022 targets, thus preserving the released modelled-residue selection and scoring complete finite backbones \citep{protein:cameo,protein:dplm2}. 

\subsection{Training and Sampling}
\label{app:protein:training}
\label{app:protein:noise}

For each residue we combine five amino-acid bits (most significant bit first) and thirteen LFQ sign bits. Sequence decoding maximises bit likelihood over the 20 canonical amino acids. All 8192 structural codes remain available. The frozen DPLM-2 decoder returns N/C$\alpha$/C/O coordinates. We train the 396.23M-parameter denoiser from scratch for 300{,}000 steps (Table~\ref{tab:protein:training}) and use the validation-selected checkpoint at step 245{,}000 for every task.

\textbf{Thirteen structural bits.} The DPLM-2 codec quantises a 13-dimensional latent vector by testing whether each of the 13 learned values is greater than zero. These bits inform each residue position with one of DPLM-2's $2^{13}=8192$ possible structural codes \citep{protein:dplm2}. CoBit preserves these codes by representing each one of them as a bit, making the conversion between a stored code and its thirteen bits exact, also preserving the codec's full vocabulary. The frozen decoder reconstructs backbone coordinates from the sequence structural codes, and for that reason thirteen bits are sufficient to encode each codec output, although this does not imply lossless encoding of arbitraty coordinates (the learned structural codec itself is lossy).

Figure~\ref{fig:protein:tokenizer} shows the 20-residue excerpt of a retained 100-residue sample as a sequence, backbone and matrix of bits. Each row represents one residue, combining five amino-acid bits with thirteen structural bits (ordered most significant bit first). The figure also separates the observed-channel choices for forward folding, inverse folding and joint generation from the sampling loop. At noise level $\sigma_k$, the same denoiser $D_\theta$ maps a noisy continuous state $Y_k$ to clean-bit probabilities $P_k$. A DDIM update produces $Y_{k+1}$, which is passed through the denoiser again at the next noise level to obtain $P_{k+1}$.

\par\addvspace{6pt}\noindent
\begin{minipage}{\linewidth}
    \makeatletter\def\@captype{figure}\makeatother
    \centering
    \includegraphics[width=\linewidth]{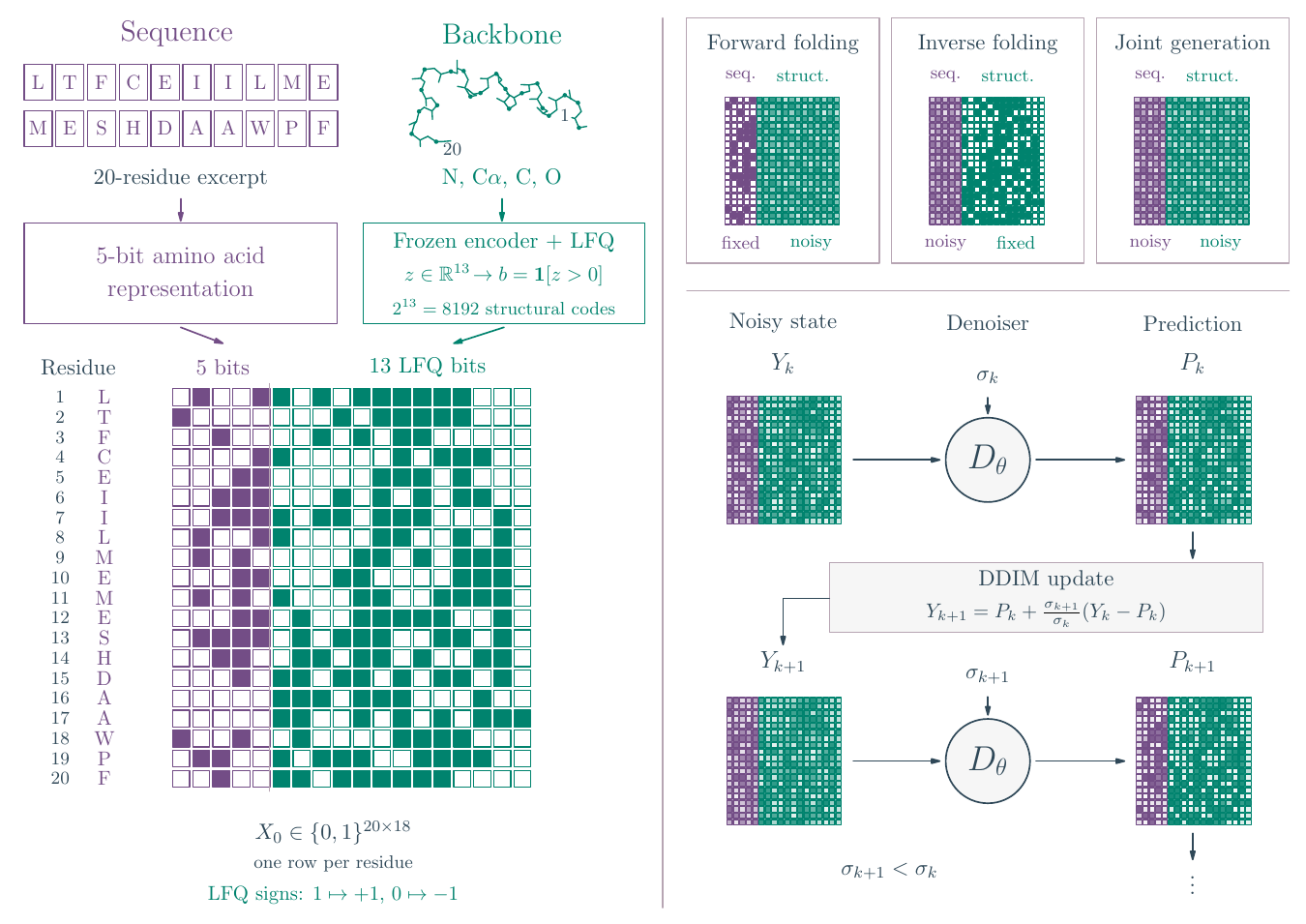}
    \caption{Protein representation and denoising. The displayed equation describes the deterministic, unguided DDIM core. 
        (a) Residue-aligned sequence and LFQ bits.
    (b) Task conditioning and successive DDIM updates with the same denoiser. }
    \label{fig:protein:tokenizer}
\end{minipage}\par\addvspace{6pt}

\par\addvspace{6pt}\noindent
    \begin{minipage}{\linewidth}
        \makeatletter\def\@captype{table}\makeatother
        \centering\small
        \caption{Training configuration of the reported protein run.}
        \label{tab:protein:training}
        \setlength{\tabcolsep}{4pt}
        \begin{tabular}{@{}ll@{}}
        \toprule
        Component & Setting \\
        \midrule
        Transformer & 20 layers; width 1024; 16 heads; FFN 4096; RoPE; SwiGLU \\
        Optimisation & AdamW; $\beta=(.9,.95)$; weight decay .01; gradient clip 1 \\
        Learning rate & $1.1\!\times\!10^{-4}$; 10k warm-up; cosine decay to $10^{-6}$ \\
        Batch / precision & Effective batch 512; accumulation 2; 8 GPUs; FP32; TF32 allowed \\
        Regularisation & Dropout .1; EMA .9999; self-conditioning probability .5 \\
        Noise / loss & Log-normal warm-up, then entropy-derived (App.~\ref{app:method}); EDM-weighted bit loss \\
        Noise / conditioning & $\sigma\in[.002,80]$; $\sigma_{\rm data}=.5$; condition dropout .1 \\
        Conditioning / seed & Length conditioning; seed 42 \\
        \bottomrule
    \end{tabular}
\end{minipage}\par\addvspace{6pt}

\textbf{Sampling configuration.} All protein results in \S\ref{sec:exp:protein} use entropic DDIM with $G=50$ noise-grid points, classifier-free guidance of 5 for conditional tasks, and no guidance for joint generation. The grid is built from the checkpoint's entropy profile (Appendix~\ref{app:method}) with weight $[\sigma^3/(\sigma^3+0.1^3)]\sqrt{e(\sigma)/\sigma^2}$. Noise levels range from 0.002 to 80 and we keep structure-noise scale=1. Stochastic sampling uses churn 10 and we keep noise multiplier fixed at 1, applied between the 10th and 90th percentiles of the entropy-based cumulative distribution.

For all protein evaluations, we use CoBit's existing \texttt{refined} self-conditioning mode (Appendix~\ref{app:method}). This mode refreshes the self-conditioning estimate at the current noisy state instead of reusing the previous prediction (which would require additional denoiser passes). The 50-point grid therefore requires 99 denoiser calls for 49 integration updates. 99 branch evaluations for joint generation and 198 for conditional generation with classifier-free guidance. These counts are derived from the sampler implementation and exclude encoding, decoding, and evaluation. Appendix~\ref{app:protein:pareto} describes how this configuration was selected and compares it with the other sampling settings

\subsection{Evaluation.}
\label{app:protein:metrics}

For inverse folding, we generate four sequences per CATH target. Amino-acid recovery (AAR) averages residue identity over candidates, then targets. Only the first sequence is refolded while forward metrics average four predictions per CAMEO target, then targets. Joint generation uses 20 pairs at each length in $\{100,200, \,...\,,500\}$, retaining all 100 draws without quality filtering or replacement. Seeds are derived deterministically from campaign seed 20260911 and sample IDs and  none of the aggregate scores use reference-based candidate selection, MSA or retrieved homologues. This cohort has 76 distinct generated sequences (repeated sequences remain in the denominator). Each joint sweep setting uses the same 100 length/seed assignments.

\textbf{One-redesign protocol.} We follow the published PMPNN1 procedure of \citet{protein:simpledesign} by redesigning each generated backbone with proteinMPNN and refold that sequence with ESMFold. We use full N/C$\alpha$/C/O backbones and temperature $.1$. Only the first ProteinMPNN sequence (candidate zero) is refolded, without selection by refolding score. ESMFold uses the Hugging Face implementation with \texttt{num\_recycles=3}, an FP16 trunk and chunk size 128 (reduced to 64 or 32 only on memory exhaustion). Aggregate scores use no reference-based candidate selection, multiple sequence alignments or retrieved homologues \citep{protein:proteinmpnn,protein:esmfold}.

\paragraph{Metrics and uncertainty.} Native structures were used to calculate forward TM, lDDT, and RMSD. Self-consistency compares refolds with conditioned or generated backbones. We report $D_{\rm TM}=100\,\Pr(\mathrm{scTM}\geq.9)$ and $D_{\rm RMSD}=100\,\Pr(\mathrm{scRMSD}\leq2\,\text{\AA})$ separately. AAR is undefined for unconditional pairs. For all joint settings, the 100-pair cohort, compiled TM-align (seed 20240303) is normalised by the generated reference length, and Kabsch RMSD uses every corresponding C$\alpha$ atom, rather than TM-align's aligned subset. All 2,000 direct/redesign refolds and 100 direct refolds pass coordinate-hash and sequence checks and rerunning TM-align and recomputing RMSD reproduces every saved score. No fallback or failed refold enters this cohort. Conditional scores retain the original TM-align/Kabsch-fallback implementation although per-score fallback use was not recorded.

Mean-score 95\% intervals use 10{,}000 bootstrap resamples: conditional targets with within-target averaging (seed 20260909), and generated pairs resampled within each length group to preserve the 20-per-length allocation (seed 20260924). Success-rate intervals use the Wilson method. Table~\ref{tab:protein:headline} gives aggregate results.

Figure~\ref{fig:protein:lengthdist} shows scTM, scRMSD and refold pLDDT for all 100 redesigned backbones, with 20 samples at each length. Coordinate-level success is higher at 100--200 residues. Meanwhile, at 500 residues, 17/20 backbones pass the strict TM criterion, but only 2/20 pass strict RMSD (Figure~\ref{fig:protein:strict}). This pattern is consistent with reduced geometric precision at longer lengths, although this evaluation does not isolate its cause. Refold pLDDT measures ESMFold confidence, not agreement with the generated backbone.

\par\addvspace{6pt}\noindent
    \begin{minipage}{\linewidth}
        \makeatletter\def\@captype{table}\makeatother
        \centering\small
        \caption{Task results under the reported sampling configuration; brackets give 95\% intervals. Joint rows use ProteinMPNN redesign.}
        \label{tab:protein:headline}
        \setlength{\tabcolsep}{4pt}
        \begin{tabular}{@{}llr@{}}
        \toprule
        Task and sample count & Metric & Estimate [95\% interval] \\
        \midrule
        Inverse: 128 targets, 512 sequences & AAR (\%) $\uparrow$ & 44.10 [40.21, 48.21] \\
        \quad First sequence: 128 refolds & scTM $\uparrow$ & .739 [.704, .770] \\
        & scRMSD (\AA) $\downarrow$ & 6.104 [5.251, 7.020] \\
        & $D_{\rm TM}$ / $D_{\rm RMSD}$ (\%) $\uparrow$ & 25.00 / 23.44 \\
        Forward: 183 targets, 732 predictions & TM $\uparrow$ & .554 [.516, .591] \\
        & lDDT $\uparrow$ & .541 [.508, .575] \\
        & RMSD (\AA) $\downarrow$ & 11.316 [10.173, 12.492] \\
        Joint: 100 backbones, one redesign each & scTM $\uparrow$ & .899 [.878, .917] \\
        & scRMSD (\AA) $\downarrow$ & 3.417 [2.817, 4.106] \\
        & $D_{\rm TM}$ (\%) $\uparrow$ & 75.00 [65.70, 82.45] \\
        & $D_{\rm RMSD}$ (\%) $\uparrow$ & 36.00 [27.27, 45.76] \\
        \bottomrule
        \end{tabular}
    \end{minipage}\par\addvspace{6pt}

\par\addvspace{6pt}\noindent
\begin{minipage}{\linewidth}
    \makeatletter\def\@captype{figure}\makeatother
    \centering
    \includegraphics[width=\linewidth]{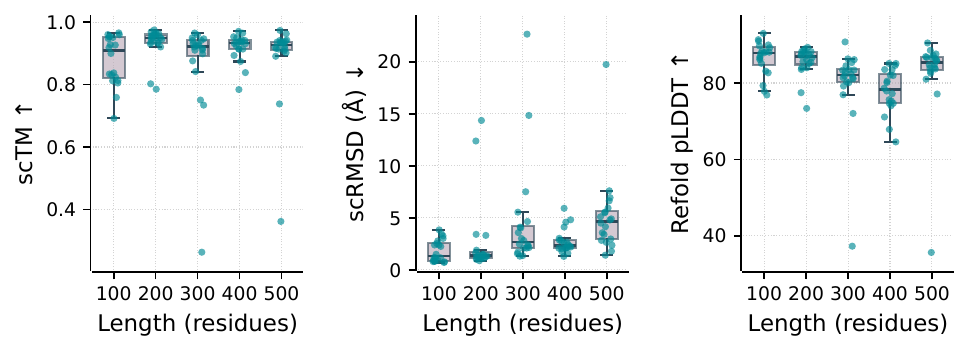}
    \caption{Backbone evaluation by length for the 100 reported samples (20 per length), each with one ProteinMPNN redesign and ESMFold refold. Left to right: scTM, scRMSD and mean C$\alpha$ pLDDT of the refold. Points show every sample and boxes show the median and interquartile range, with whiskers extending to the most extreme values within 1.5 interquartile ranges. pLDDT is predicted refold confidence.}
    \label{fig:protein:lengthdist}
\end{minipage}\par\addvspace{6pt}

\subsection{Direct consistency of reported pairs}
\label{app:protein:direct}

We refold CoBit's own sequences for the same 100 pairs reported in \S\ref{sec:exp:protein}, using the identical ESMFold, TM-align and all-C$\alpha$ RMSD settings as the one-redesign evaluation. This measures consistency of the generated sequence--structure pairs directly. Both evaluations retain all 20 samples at each length.

\par\addvspace{6pt}\noindent
    \begin{minipage}{\linewidth}
    \makeatletter\def\@captype{table}\makeatother
    \centering\small
    \caption{Direct refolding of CoBit's sequences for the 100 pairs reported in the main text. Success rates are percentages.}
    \label{tab:protein:direct}
    \setlength{\tabcolsep}{4pt}
    \begin{tabular}{@{}lrrrrr@{}}
    \toprule
    Refolded sequence & $n$ & scTM $\uparrow$ & scRMSD (\AA) $\downarrow$ & $D_{\rm TM}$ & $D_{\rm RMSD}$ \\
    \midrule
    CoBit & 100 & .911 & 3.060 & 69.0 & 41.0 \\
    \bottomrule
    \end{tabular}
\end{minipage}\par\addvspace{6pt}

The 95\% scTM interval is [.894, .926] and the scRMSD interval is [2.594, 3.644]\,\AA{} under the length-stratified bootstrap. Wilson intervals are [59.4, 77.2]\% for $D_{\rm TM}$ and [31.9, 50.8]\% for $D_{\rm RMSD}$. Direct and redesigned sequences share the same scoring protocol, but their results measure different aspects of consistency and neither dominates on both strict criteria.

\subsubsection{Sampler selection and cost}
\label{app:protein:pareto}

We sweep CFG $\in\{0,1,2,3,4,5\}$, $G\in\{50,100,150,200\}$ and Karras fraction $\alpha\in\{0,.1,.2,.5,1\}$, with $\sigma_i=(1-\alpha)\sigma_i^{\rm entropic}+\alpha\sigma_i^{\rm Karras}$. The reported configuration was selected on CATH validation from the 16 pure-entropic settings with CFG 2--5. After requiring full target coverage and exact clamping, settings are ranked by maximising success (scRMSD $<2$\,\AA{} or scTM $>.5$), scTM and pLDDT, then minimising scRMSD, maximising diversity and AAR, and preferring fewer steps. 

We observed that at $G=50$, increasing entropic CFG from 0 to 5 raises inverse scTM from .708 to .739 while diversity (one minus mean pairwise identity among four designs) falls from .392 to .145. Strict success does not improve consistently (Figure~\ref{fig:protein:cfg-evaluation}). The matched 100-pair joint sweep reaches its highest mean scTM of $.907$ at  $\alpha=.1$, $G=50$, versus $.899$ for the reported $\alpha=0$, $G=50$ configuration. Within that sweep, scTM rises from .899 at the reported configuration to .903 at $\alpha=.1$, $G=200$, which costs roughly four times as many denoiser evaluations (Figure~\ref{fig:protein:pareto}). All main results use the fixed 50-point entropic configuration.

\par\addvspace{6pt}\noindent
\begin{minipage}{\linewidth}
    \makeatletter\def\@captype{figure}\makeatother
    \centering
    \includegraphics[width=\linewidth]{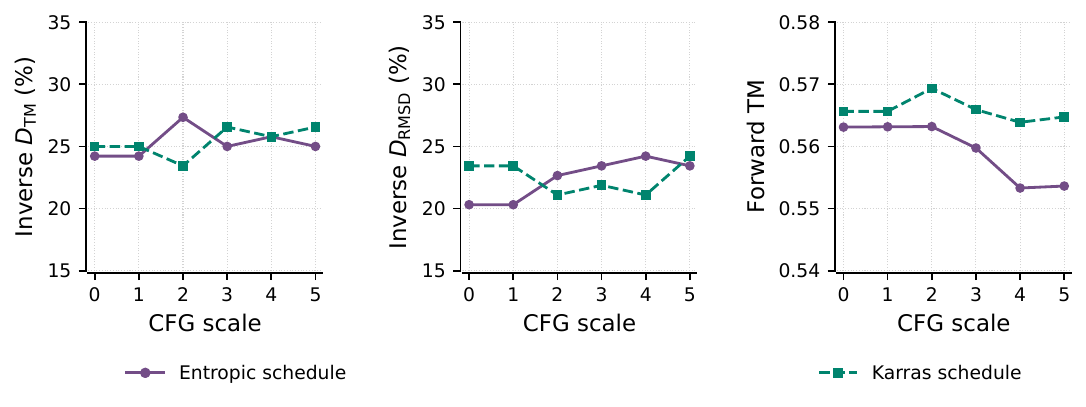}
    \caption{CFG sweep at $G=50$: inverse success rates on 128 targets and mean
    forward TM on 183 targets. Curves show point estimates. CFG 0 costs 99 branch
    evaluations; positive CFG costs 198.}
    \label{fig:protein:cfg-evaluation}
\end{minipage}\par\addvspace{6pt}

\par\addvspace{6pt}\noindent
\begin{minipage}{\linewidth}
    \makeatletter\def\@captype{figure}\makeatother
    \centering
   \includegraphics[width=\linewidth]{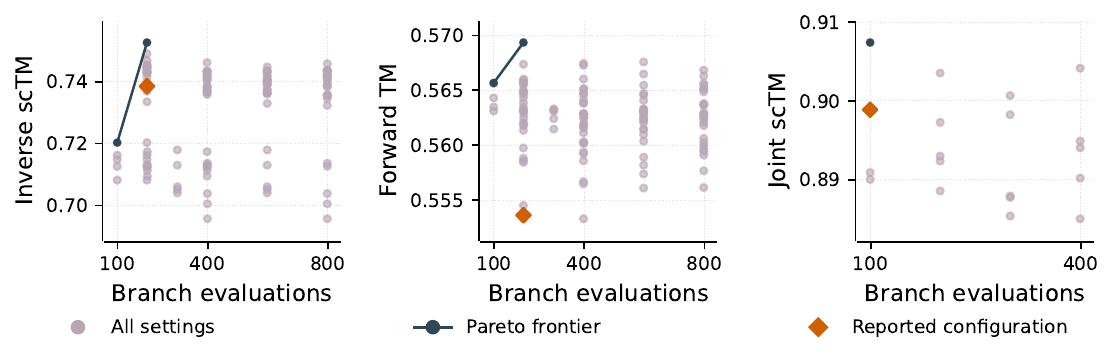}
    \caption{Quality--cost trade-offs across all settings. Lines join non-dominated point estimates. Diamonds mark the reported configuration. Each joint setting uses 100 pairs with matched length/seed assignments and one ProteinMPNN redesign per backbone. Conditional target sets are unchanged.}
    \label{fig:protein:pareto}
\end{minipage}\par\addvspace{6pt}

\subsection{Comparison context and limits}
\label{app:protein:comparison}

We present table~\ref{tab:protein:baselines} with the PMPNN1 column of \citet{protein:simpledesign}. We match its sample count ($n=100$), length ladder (100--500), one-candidate redesign/refolding procedure, strict thresholds, TM-align scoring and all-C$\alpha$ RMSD definition. In that study, it was not possible to find the exact allocation across lengths, ProteinMPNN sampling temperature and evaluator checkpoint, ESMFold recycling count and evaluator versions, or which TM-align length normalisation is retained. 

Considering the best-perfoming model in Table~\ref{tab:protein:baselines}, MultiFlow's designability distillation raises its one-redesign designability from 71\% to 88\% in its own ablation \citep{campbell2024multiflow}. CoBit does not leverage distillation and uses unpaired sequence replay without a direct geometric loss. Cross-study comparisons therefore do not isolate representation or training effects, and computational consistency alone does not establish biological function.
\endgroup

\end{document}